\documentclass[lettersize,journal]{IEEEtran}
\usepackage{booktabs}
\usepackage{diagbox}
\usepackage{makecell}
\usepackage{array}
\usepackage{graphicx,amssymb,amsmath}
\usepackage{multicol}
\usepackage[noadjust]{cite} 
\usepackage[justification=centering]{caption}
\usepackage[font=small]{caption}
\usepackage{setspace}
\usepackage{subfigure}
\usepackage{graphicx}
\usepackage{float}
\usepackage {url}
\usepackage{stfloats}
\usepackage[normal]{threeparttable}
\usepackage{amsthm,pifont}
\usepackage{flushend}
\usepackage{cases,subeqnarray}
\usepackage{bm,multirow,bigstrut}
\usepackage{amsmath, amsthm, amssymb}
\usepackage{textcomp}
\usepackage{latexsym,bm}
\usepackage{booktabs}
\usepackage{mathtools}
\usepackage{dsfont}
\usepackage{extarrows}
\usepackage{epsfig}
\usepackage{epsfig}
\usepackage{epstopdf}
\usepackage[table]{xcolor}
\usepackage{colortbl} 
\usepackage{multirow, hhline}
\usepackage{xcolor}
\usepackage{array} 
\usepackage{hyperref}
\usepackage{tikz}
\usepackage{makecell}
\usepackage{enumitem}
\usepackage{comment}
\usepackage{pifont}
\usepackage{graphicx}
\usepackage{tabularray}
\usetikzlibrary{shapes.geometric, arrows, positioning}
    
\theoremstyle{plain}

\theoremstyle{plain}

\usepackage{amsmath}

\usepackage[edges]{forest}
\usetikzlibrary{shadows}

\usepackage[ruled,linesnumbered]{algorithm2e}

\usetikzlibrary{shapes.geometric, arrows}

\definecolor{lightblue}{rgb}{0.68, 0.85, 0.90}
\definecolor{lightgreen}{rgb}{0.56, 0.93, 0.56}
\definecolor{lightpurple}{rgb}{0.75, 0.58, 0.92}

\tikzstyle{box} = [rectangle, rounded corners, minimum width=3cm, minimum height=1cm,text centered, draw=black, fill=lightblue]
\tikzstyle{boxgreen} = [rectangle, rounded corners, minimum width=3cm, minimum height=1cm,text centered, draw=black, fill=lightgreen]
\tikzstyle{boxpurple} = [rectangle, rounded corners, minimum width=3cm, minimum height=1cm,text centered, draw=black, fill=lightpurple]
\tikzstyle{line} = [draw, -latex]

\IEEEoverridecommandlockouts
\begin{document}
\title{ Leveraging Large Vision Model for Multi-UAV Co-perception in Low-Altitude Wireless Networks }

\author{
Yunting Xu, Jiacheng Wang, Ruichen Zhang, Changyuan Zhao, Yinqiu Liu, Dusit Niyato,~\IEEEmembership{Fellow,~IEEE}, \\ Liang Yu,  Haibo Zhou,~\IEEEmembership{Fellow,~IEEE}, and Dong In Kim,~\IEEEmembership{Life Fellow,~IEEE} 

\thanks{Y. Xu, J. Wang, R. Zhang, C. Zhao, Y. Liu, and D. Niyato are  with the College of Computing and Data Science, Nanyang Technological University, Singapore, 639798 (e-mail: yunting.xu@ntu.edu.sg, jiacheng.wang@ntu.edu.sg, ruichen.zhang@ntu.edu.sg, zhao0441@e.ntu.edu.sg, yinqiu001@e.ntu.edu.sg, dniyato@ntu.edu.sg).}
\thanks{L. Yu is with the Alibaba Cloud, Hangzhou, China, 311121 (e-mail: liangyu.yl@alibaba-inc.com).} 
\thanks{H. Zhou is with the School of Electronic Science and Engineering, Nanjing University, Nanjing, China, 210023 (e-mail: haibozhou@nju.edu.cn).}
\thanks{D. I. Kim is with the Department of Electrical and Computer Engineering, Sungkyunkwan University, Suwon 16419, South Korea (email: dongin@skku.edu).}
}

\maketitle
\begin{abstract}

Multi-uncrewed aerial vehicle (UAV) cooperative perception has emerged as a promising paradigm for diverse low-altitude economy applications, where complementary multi-view observations are leveraged to enhance perception performance via wireless communications. However, the massive visual data generated by multiple UAVs poses significant challenges in terms of communication latency and resource efficiency. To address these challenges, this paper proposes a communication-efficient cooperative perception framework, termed Base-Station-Helped UAV (BHU), which reduces communication overhead while enhancing perception performance. Specifically, we employ a Top-K selection mechanism to identify the most informative pixels from UAV-captured RGB images, enabling sparsified visual transmission with reduced data volume and latency. The sparsified images are transmitted to a ground server via multi-user MIMO (MU-MIMO), where a Swin-large-based MaskDINO encoder extracts bird’s-eye-view (BEV) features and performs cooperative feature fusion for ground vehicle perception. Furthermore, we develop a diffusion model–based deep reinforcement learning (DRL) algorithm to jointly select cooperative UAVs, sparsification ratios, and precoding matrices, achieving a balance between communication efficiency and perception utility. Simulation results on the Air-Co-Pred dataset demonstrate that, compared with traditional CNN-based BEV fusion baselines, the proposed BHU framework improves perception performance by over 5\% while reducing communication overhead by 85\%, providing an effective solution for multi-UAV cooperative perception under resource-constrained wireless environments.







\end{abstract}

\IEEEpeerreviewmaketitle

\section{Introduction}

Multi-unmanned aerial vehicle (UAV) cooperative perception has recently emerged as a promising technology for a variety of low-altitude economy applications, such as intelligent transportation, urban surveillance, and autonomous inspection \cite{10834514, wang2023secure, xu2021leveraging}. By exploiting the complementary multi-view observations from spatially distributed UAVs, cooperative perception systems can substantially enhance perception accuracy and robustness compared with single-UAV solutions \cite{xu2021throughput, 11283103}. However, realizing such performance gains in practical wireless environments remains challenging due to the massive volume of visual data generated by multiple UAVs \cite{kang2022task, xue2024cooperative}. Additionally, the heterogeneous and time-varying channel conditions between UAVs and base stations (BSs), together with the limited wireless time–frequency resources, make it difficult to satisfy the stringent latency requirements for real-time UAV communications, thereby hindering the efficient data transmission for cooperative multi-UAV perception \cite{wang2025exploring}.

To reduce communication overhead between the cooperative UAVs, instead of transmitting raw UAV images or dense visual features, recent studies have explored feature-level compression or sparsification to alleviate communication bottlenecks \cite{hu2022where2comm, zhao2}. By dynamically adjusting the transmission data size of visual images, these approaches can scale effectively in bandwidth-limited and latency-sensitive scenarios \cite{wang2020v2vnet, hu2024communication}. 
Nevertheless, since UAVs operate as resource-constrained edge devices with limited energy and computational capacity, they are typically equipped with lightweight network modules for onboard image processing \cite{singh2024communication}. As a result, compression or sparsification approaches often suffer from limited reconstruction and feature extraction capabilities, which significantly degrades the quality of perception.
Therefore, effectively balancing perception performance and communication efficiency remains a challenging problem, particularly for multi-UAV cooperative perception tasks.

Recent development of large vision models (LVMs), such as Vision Transformers (ViT) \cite{dosovitskiy2020image} and Self-distillation with No Label (DINO) \cite{caron2021emerging}, have demonstrated advanced representation and generalization capabilities compared with traditional Convolutional Neural Networks (CNNs). Based on the self-attention mechanism, LVMs can effectively capture long-range dependencies and semantic relationships to extract rich visual information, thus achieving state-of-the-art performance across diverse image tasks such as object classification, instance detection, and semantic segmentation \cite{xu2025integrating}. As a consequence, LVMs have shown substantial potential for enhancing multi-UAV cooperative perception, enabling more accurate image reconstruction, feature extraction, and scene understanding from information-reduced visual data \cite{simeoni2025dinov3}. 
Nevertheless, while LVMs are capable of achieving high perception performance, their large-scale model parameters and substantial computational overhead pose significant challenges to the practical implementation in real-world UAV applications.

Motivated by the growing demand to fully exploit the potential of LVMs, this paper proposes a communication-efficient multi-UAV perception framework, named Base-Station-Helped UAV (BHU), aiming to reduce the communication overhead while enhancing perception performance. The core idea of the BHU framework is to transmit sparsified UAV images to a ground server via wireless communication, where an LVM–based encoder extracts bird’s-eye-view (BEV) features from the aerial images and performs cooperative feature fusion for downstream perception tasks.
Specifically, we model the communication between cooperative UAVs and the BS as a multi-user multiple-input multiple-output (MU-MIMO) system. To further achieve an effective trade-off between perception utility and communication efficiency, we adaptively select the cooperative UAVs, image sparsification ratios, precoding strategies for each UAV based on the channel state information, establishing a communication-aware multi-UAV cooperative perception mechanism for low-altitude economy scenarios. The main contributions are summarized as follows:

\begin{itemize}

\item 
We introduce an importance score–based Top-K mechanism for sparsified visual transmission, which selects the most informative pixels from UAV-captured images and substantially reduces both transmission data volume and communication latency.

\item 
We leverage a Swin-large-based MaskDINO encoder \cite{li2023mask} to extract BEV features and perform cooperative feature fusion for vehicle instance and motion-aware perception, thereby significantly enhancing perception performance through the strong representation capability of LVMs.

\item 
We propose a diffusion model–based deep reinforcement learning (DRL) framework to jointly optimize cooperative UAV selection, Top-K sparsification ratios, and precoding matrices. A denoising diffusion implicit model (DDIM) is integrated  to improve the training stability and efficiency of the DRL framework.  

\item Extensive simulations conducted on Air-Co-Pred dataset \cite{wang2024drones} demonstrate that the proposed BHU framework enhances perception utility by over 5\% and reduces communication overhead by 85\%, offering a communication-efficient solution for multi-UAV cooperative perception in resource-constrained wireless environment.

\end{itemize}

The remainder of this paper is organized as follows. Section II reviews the related work. Section III introduces the system model of multi-UAV cooperative perception. Section IV describes the proposed BHU framework. Section V presents the DDIM-based DRL solution. Simulation results are provided in Section VI. Finally, Section VII concludes this paper.

\section{Related Work}

\subsection{Visual Data Reduction and UAV Wireless Transmission}

The transmission of high-resolution visual data from UAVs to ground infrastructure poses a major challenge due to stringent bandwidth and latency constraints \cite{zhang2025robust, zhao1}. To mitigate communication overhead, extensive research has investigated image and feature compression techniques for efficient transmission \cite{xue2024sparse}. For example, Giordano \emph{et al.} \cite{giordano2017roi} developed an onboard region-of-interest (RoI)-based compression framework for remote sensing images, combining RoI identification with differential compression schemes to reduce transmission volume while preserving target quality. Additionally, sparsification methods adopt token selection mechanisms to transmit only informative content, such as the spatial confidence–guided sparse messaging scheme proposed in \emph{Where2comm} \cite{hu2022where2comm} and the codebook-based discrete message representation employed by \emph{CodeFilling} \cite{hu2024communication}, thereby enabling communication-efficient collaborative perception.

Except for image processing, advanced wireless technologies have been widely adopted to enhance transmission performance in UAV communications \cite{yunting1, zhang2025ris, TNSE_ISAC}. For instance, Geraci \emph{et al.} \cite{8528463} evaluated the achievable throughput of UAV cellular communications within massive MIMO systems, demonstrating that the adoption of massive MIMO technology enables approximately a 50\% increase in the number of UAVs that meet the minimum SINR threshold compared with conventional cellular deployments. In the context of multi-UAV scenarios, Zeng \emph{et al.} \cite{10309145} investigated the optimization of MU-MIMO beamforming matrices based on the weighted minimum mean square error (WMMSE) method. Simulation results validated that the proposed MU-MIMO beamforming optimization scheme can effectively increase the sum rate of all UAV users by approximately 30\%. Furthermore, Elwekeil \emph{et al.} \cite{9973340} introduced cell-free massive MIMO into UAV communications, where effective interference management is achieved through power control mechanism. By transforming the inter-UAV interferences into useful signal components, cell-free mechanism significantly enhances spectral efficiency and improves the per-user communication rate performance compared to traditional multi-cell network deployment \cite{yukai2}.

\subsection{Cooperative Perception for Multi-UAV Systems}
Cooperative perception aggregates complementary observations from multiple UAVs to enhance detection performance, yet it inevitably incurs additional coordination and communication overhead \cite{xu2025mixture}. Existing cooperative perception frameworks can be broadly categorized into UAV-centric and infrastructure-assisted approaches. In UAV-centric cooperation, UAVs exchange perception features or intermediate results directly with neighboring UAVs, enabling distributed collaboration without relying on ground infrastructures \cite{feng2024u2udata}. For instance, Liu \emph{et al.} presented \emph{When2com} \cite{liu2020when2com}, where one UAV agent is required to perform a perception task and share information with other agents. Based on communication group construction, \emph{When2com} significantly reduces communication bandwidth while maintaining superior performance. In infrastructure-assisted cooperation, a ground station or edge server aggregates information from multiple UAVs for centralized processing. A representative multi-drone collaborative framework is \emph{UAV-CodeAgents} \cite{sautenkov2025uav}, which employs an Airspace Management Agent to provide centralized reasoning capabilities for coordinating UAV agents. Simulation demonstrates that the proposed system can effectively enhance the reliability in coordinating UAV operations, achieving a 93\% success rate in a cooperative fire detection task.

\subsection{LVMs for UAV-based Visual Perception}

Recent advances in LVMs, such as ViT, Swin Transformer \cite{liu2021swin}, self-supervised DINO \cite{caron2021emerging}, and unified detection and segmentation frameworks have demonstrated remarkable representation capability and generalization performance \cite{xu2025enhancing, 10736570}. Specifically, the ViTs pioneered the patch-based Transformer backbones in visual recognition, exhibiting remarkable scalability under large-scale pretraining. Hierarchical Transformer backbones such as Swin Transformer improve efficiency via shifted-window attention while retaining strong dense-prediction performance. DINO employs knowledge distillation and contrastive learning techniques to acquire high-quality representations without human annotations, revealing rich semantic structures in the learned representations.
Building upon end-to-end Detection Transformer (DETR) \cite{carion2020end} paradigms, MaskDINO \cite{li2023mask}, extending DINO with a mask prediction branch to support instance, panoptic, and semantic segmentation in a unified Transformer-based framework.

Motivated by these advanced capabilities, recent research has explored leveraging LVMs for a variety of UAV perception tasks. For aerial object detection, Zhang \emph{et al.} proposed UAV-DETR \cite{zhang2025uav}, which introduced a multi-scale feature fusion mechanism to capture both spatial and frequency information, thereby significantly enhancing the detection performance. In the remote sensing tasks, Wang \emph{et al.} \cite{wang2022novel} adopted Swin Transformer as the backbone and designed a densely connected feature aggregation decoder for fine-resolution semantic segmentation.
Beyond task-specific applications, Shan \emph{et al.} \cite{shan2025ros} built ROS-SAM, a redesigned data processing pipeline based on the Segment Anything Model (SAM) \cite{kirillov2023segment} and a mask decoder to obtain high-quality interactive segmentation for remote sensing moving objects. Nonetheless, existing LVM-based UAV perception studies primarily focus on perception accuracy or
inference efficiency, while the joint optimization among transmission efficiency collaboration policies, and perception utility in multi-UAV settings remains explored.

\begin{figure}
\centering
\includegraphics[width=0.49\textwidth]{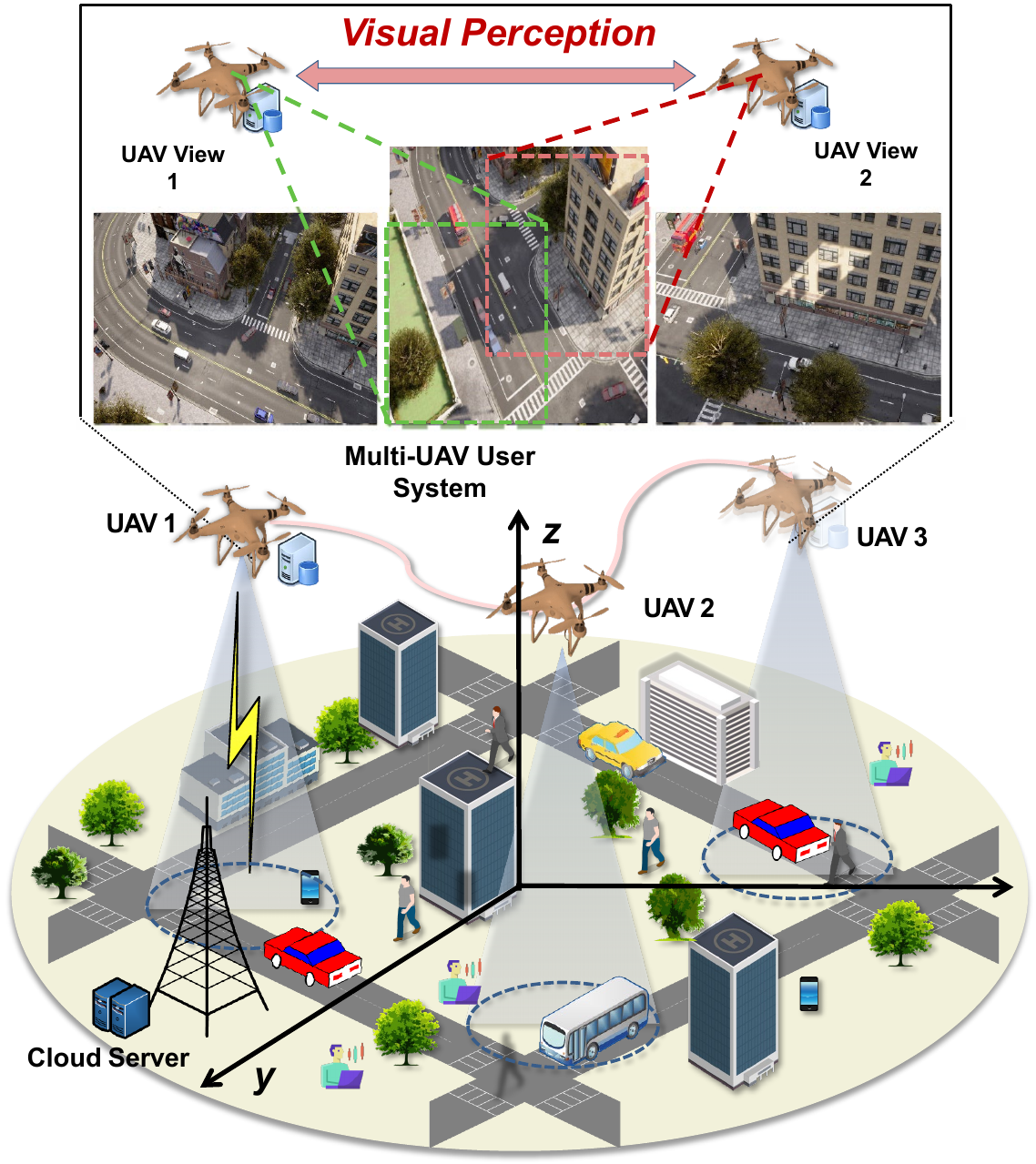} 
\centering
\captionsetup{justification=justified,format=plain}
\caption{ Illustration of multi-UAV cooperative perception scenario. The UAV-captured aerial images are transmitted to a ground server via MU-MIMO communication, where multi-view information is fused to enable cooperative feature learning for vehicle instance and motion detection in low-altitude economy scenarios. }
\label{figure_system_model}
\end{figure}

\section{System Model}

\subsection{3D Geometry-Based Wideband Channel}

In this paper, we consider a multi-UAV cooperative perception scenario as illustrated in Fig.~\ref{figure_system_model}. Define the set of UAV as $\mathcal{U} = \{1, \ldots, U \}$ and the set of BS as $\mathcal{B} = \{1, \ldots, B \}$. For a MIMO transmission system, the UAV and the ground BS are equipped with $t_x$ transmit antennas and $r_x$ receive antennas, respectively. For the propagation channel between the $u$-th UAV user and the $b$-th BS, the $\ell$-th path component $\mathbf{H}^{\ell}_{u,b} \in \mathbb{C}^{r_x\times t_x}$ is characterized by the following geometry-based 3D wideband model \cite{3GPP38.901} 
\begin{equation}\label{3D channel model}
\begin{split}
\mathbf{H}^{\ell}_{u,b}(t, f)= & \rho_{u, b, \ell} e^{j 2 \pi \nu_{u, b, \ell} t} e^{-j 2 \pi f \tau_{u, b, \ell}} \mathbf{a}_{r}\left(\theta_{u, b, \ell}^{a z}, \theta_{u, b, \ell}^{e l}\right) \\ 
& \cdot \mathbf{a}_{t}\left(\phi_{u, b, \ell}^{a z}, \phi_{u, b, \ell}^{e l}\right)^H, \quad \forall u \in \mathcal{U}, ~ \forall b \in \mathcal{B},
\end{split}
\end{equation}
where $t$ and $f$ denote the continuous variables in the time and frequency domains. For the $\ell$-th channel path, $\rho_{u, b, \ell}$ and $\tau_{u, b, \ell}$ are the complex path gain and propagation delay, respectively, while $\nu_{u, b, \ell}$ denotes the Doppler shift. $\mathbf{a}_{r}(\theta_{u, b, \ell}^{a z}, \theta_{u, b, \ell}^{e l})$ and $\mathbf{a}_{t}(\phi_{u, b, \ell}^{a z}, \phi_{u, b, \ell}^{e l})$ represent the 3D array response vectors at the $b$-th BS and the $u$-th UAV, parameterized by the azimuth and elevation angles of arrival (AoA) $\theta_{u, b, \ell}^{az/el}$ and the angles of departure (AoD) $\phi_{u, b, \ell}^{az/el} $, respectively. Assume the BS uses a uniform planar array (UPA) with $(N_x\times N_y)$ antenna elements, the antenna array response can be expressed as: 
\begin{equation}\label{BS Impulse Response}
\begin{split}
\mathbf{a}_{r}(\theta_{u, b, \ell}^{a z}, \theta_{u, b, \ell}^{e l})  =
\mathbf{a}_{x} (\theta_{u, b, \ell}^{a z}, \theta_{u, b, \ell}^{e l} )  \otimes \mathbf{a}_{y} (\theta_{u, b, \ell}^{a z}, \theta_{u, b, \ell}^{e l} ) ,
\end{split}
\end{equation}
where $\mathbf{a}_x\left(\cdot\right)$ and $\mathbf{a}_y\left(\cdot\right)$ are the impulse response vectors of the horizontal and vertical antenna arrays, and $\otimes$ represents the Kronecker product operation. Denote the inter-element spacing as $d$ and the carrier wavelength as $\lambda_c$, $\mathbf{a}_x\left(\cdot\right)$ and $\mathbf{a}_y\left(\cdot\right)$ can be formulated as:
\begin{equation}\label{x Impulse Response}
\begin{split}
\mathbf{a}_{x} ( \theta_{u, b, \ell}^{a z}, &  \theta_{u, b, \ell}^{e l}  )= \frac{1}{\sqrt{N_x}}\left[1, e^{j 2\pi \frac{d}{\lambda_{c}} \sin \left( \theta_{u, b, \ell}^{e l} \right) \cos \left( \theta_{u, b, \ell}^{a z} \right)}, \right. \\
& \quad ~\left. ~\cdots,~ e^{j 2\pi \frac{d}{\lambda_{c}}\left(N_1-1\right) \sin \left( \theta_{u, b, \ell}^{e l} \right) \cos \left( \theta_{u, b, \ell}^{a z}\right)} \right]^{\mathrm{T}},
\end{split}
\end{equation}
\begin{equation}\label{y Impulse Response}
\begin{split}
 \mathbf{a}_{y} ( \theta_{u, b, \ell}^{a z}, & \theta_{u, b, \ell}^{e l} ) = \frac{1}{\sqrt{N_y}}\left[1, e^{j 2\pi \frac{d}{\lambda_{c}} \sin \left( \theta_{u, b, \ell}^{e l} \right) \sin \left(\theta_{u, b, \ell}^{a z} \right)}, \right. \\
& \quad ~ \left. ~\cdots, ~ e^{j 2\pi \frac{d}{\lambda_{c}} \left(N_2-1\right) \sin \left( \theta_{u, b, \ell}^{e l} \right) \sin \left( \theta_{u, b, \ell}^{a z} \right)} \right]^{\mathrm{T}}.
\end{split}
\end{equation}

Similarly, the UAV transmit array response $\mathbf{a}_{t}(\phi_{u, b, \ell}^{a z}, \phi_{u, b, \ell}^{e l})$ is obtained  following the same procedure as $\mathbf{a}_{r}\left(\cdot\right)$.
To explicitly model the LoS component in aerial communication links, Rician channel model is employed to characterize the relative contributions of the LoS and NLoS components, given by
\begin{equation}\label{Rician model}
\begin{split}
\mathbf{H}_{u,b}(t, f)= \sqrt{\frac{\text{k}_r}{\text{k}_r+1}} \mathbf{H}^{LoS}_{u,b}(t, f) + \sqrt{\frac{1}{\text{k}_r+1}} \sum_{\ell=1}^{L} \mathbf{H}^{\ell}_{u,b}(t, f),
\end{split}
\end{equation}
where $\text{k}_r$ denotes the Rician K-factor. $\mathbf{H}^{LoS}_{u,b}$ represents the single dominant $LoS$ path, while the $NLoS$ channel component is formed by $L$ scattered paths with corresponding AoA/AoD and propagation delays.

\subsection{Multi-User MIMO Transmission}

For the scenario of multi-UAV users and ground BS communications, we consider an uplink MU-MIMO OFDM transmission where $U$ UAV users simultaneously transmit signals to a multi-antenna BS over the same time–frequency resources. A binary variable ${a}_{u, b}$ is adopted to indicate the user association between UAV $u$ and BS $b$, which is expressed as:
\begin{equation}\label{lowercase_x}
\begin{split}
{a}_{u, b}= \begin{cases}1 & \text { if UAV}~ u ~\text {accesses BS}~b, \\ 0 & \text { otherwise.} \end{cases}
\end{split}
\end{equation}

Consequently, the signal received at BS $b$ is given by 
\begin{equation}\label{received signal}
\begin{split}
\mathbf{y}_{b}[k][s] = \sum_{u=1}^{U} {a}_{u, b} \mathbf{H}_{u,b}[k][s]  \mathbf{W}_{u} \sqrt{p_u} x_{u}[k][s] + \mathbf{n}_{b},
\end{split}
\end{equation}
where $k \in \mathcal{K} = \{1, \cdots, K \}$ and $s \in \mathcal{S} = \{1, \cdots, S \}$ denote the indexes of the subcarrier and the OFDM symbol, respectively. $p_u$ is the transmit power of UAV $u$, while $\mathbf{W}_{u} \in \mathbb{C}^{t_x \times 1}$ denotes the precoding vector for single-stream transmission. $x_{u}[k][s]$ is the transmitted symbol with $\mathbb{E}\{|x_{u}[k][s]|^2\} =1$ and $\mathbf{H}_{u,b}[k][s] \in \mathbb{C}^{r_x \times t_x}$ represents the discrete frequency-domain MIMO channel. $\mathbf{n}_{b} \in \mathbb{C}^{{r_x} \times 1} \sim \mathcal{C N}\left(0, \sigma^2 \boldsymbol{I}_{r_x} \right) $ is the additive white Gaussian noise with the noise variance $\sigma^2$. 

Considering the perception-oriented task where the hovering UAVs maintain low mobility, the Doppler spread is assumed to be sufficiently small such that the inter-carrier interference is negligible. Let 
\begin{equation}\label{uppercase_a}
\begin{split}
\mathbf{A}_{u,b} = {a}_{u, b}\mathbf{I}_{r_x}, ~\forall u \in \mathcal{U}, ~ b \in \mathcal{B},
\end{split}
\end{equation}
and omit the indexes $k$ and $s$ for simplicity of expression, Eq. \eqref{received signal} can be transformed into
\begin{equation}\label{received signal 1}
\begin{split}
\mathbf{y}_{b}= \underbrace{ \left[ \mathbf{A}_{1,b} \mathbf{H}_{1,b}, \cdots, \mathbf{A}_{U,b} \mathbf{H}_{U,b} \right]}_{\in \mathbb{C}^{r_x \times (U \cdot t_x)}} \underbrace{\left[\begin{array}{c}
\mathbf{W}_1 \sqrt{p_1} x_{1} \\
\vdots \\
\mathbf{W}_U \sqrt{p_U} x_{U}
\end{array} \right] }_{\in \mathbb{C}^{(U \cdot t_x) \times 1}}+ \mathbf{n}_{b}.
\end{split}
\end{equation}

Let the aggregated matrices $\mathbf{W}_{b} \in \mathbb{C}^{ (U \cdot t_x) \times U}$, $\mathbf{H}_{b} \in \mathbb{C}^{(U \cdot r_x) \times (U \cdot t_x)}$, and $\mathbf{A}_{b} \in \{0,1\}^{r_x \times (U \cdot r_x)}$ be defined as
\begin{align}\label{transmit symbol vector}
& \mathbf{W}_{b} = \operatorname{diag} \left( \sqrt{p_1} \mathbf{W}_1, ~\cdots,~ \sqrt{p_U} \mathbf{W}_U \right),\\
& \mathbf{H}_{b} = \operatorname{diag}\left( \mathbf{H}_{1,b}, ~\cdots,~ \mathbf{H}_{U,b} \right),\\
& \mathbf{A}_{b} = \left[ \mathbf{A}_{1,b}, ~\cdots,~ \mathbf{A}_{U,b} \right].
\end{align}

Eq.~\eqref{received signal 1} can be reformulated as
\begin{equation}\label{received signal 2}
\begin{split}
\mathbf{y}_{b} =  \underbrace{\mathbf{A}_{b} \mathbf{H}_{b} \mathbf{W}_{b}}_{\mathbf{H}_{\text{eff}} \in \mathbb{C}^{r_x \times U}} \mathbf{x}_{b} + \mathbf{n}_{b},
\end{split}
\end{equation}
where $ \mathbf{x}_{b} = \left[ {x}_{1,b}, ~\cdots,~ {x}_{U,b}\right]^{\mathrm{T}} \in \mathbb{C}^{U \times 1}$ denotes the aggregated vector of the transmit signals. Define $\mathbf{A}_{b} \mathbf{H}_{b} \mathbf{W}_{b}$ as the effective channel matrix $\mathbf{H}_{\text{eff}}$. Then, the received signal can be processed by a minimum mean square error (MMSE) equalizer to mitigate channel distortion \cite{yukai1}, expressed as
\begin{equation}\label{MMSE}
\begin{split}
&\mathbf{G}_{\mathrm{MMSE}}  =\left(\mathbf{H}_{\mathrm{eff}}^{\mathrm{H}} \mathbf{H}_{\mathrm{eff}}+\sigma^2 \mathbf{I}\right)^{-1} \mathbf{H}_{\mathrm{eff}}^{\mathrm{H}}.
\end{split}
\end{equation}

Consequently, denote the MMSE equalizer employed at BS $b$ as $\mathbf{G}_b \in \mathbb{C}^{U \times r_x}$, which is given by
\begin{equation}\label{MMSE q}
\begin{split}
&\mathbf{G}_{b} = \left( \left(\mathbf{A}_{b} \mathbf{H}_{b} \mathbf{W}_{b} \right)^{\mathrm{H}}\left(\mathbf{A}_{b} \mathbf{H}_{b} \mathbf{W}_{b} \right) +\sigma^2 \mathbf{I} \right)^{-1} 
\left(\mathbf{A}_{b} \mathbf{H}_{b} \mathbf{W}_{b} \right)^{\mathrm{H}}.
\end{split}
\end{equation}

We can obtain the linear estimation of the transmitted symbol vector $\widetilde{\mathbf{x}}_{b} \in \mathbb{C}^{U \times 1}$ as
\begin{equation}\label{post-equalization vector}
\begin{split}
\widetilde{\mathbf{x}}_{b} &=\mathbf{G}_{b} \mathbf{y}_{b} \\
&=\mathbf{G}_{b}  \left( \mathbf{A}_{b} \mathbf{H}_{b} \mathbf{W}_{b} \mathbf{x}_{b} + \mathbf{n}_{b} \right) \\
&=\underbrace{\mathbf{G}_{b}  \mathbf{A}_{b} \mathbf{H}_{b} \mathbf{W}_{b} }_{ \mathbf{G}_{b} \mathbf{H}_{\mathrm{eff}} \in \mathbb{C}^{U \times U}} \mathbf{x}_{b} +\underbrace{ \mathbf{G}_{b} \mathbf{n}_{b} }_{ \in \mathbb{C}^{U \times 1}}.
\end{split}
\end{equation}

Let $\mathbf{g}_u^{\mathrm{H}}$ denote the $u$-th row of $\mathbf{G}_{b}$ and $\mathbf{h}_u$ denote the $u$-th column of $\mathbf{H}_{\mathrm{eff}}$. Then, the estimated symbol $\hat{x}_u$ corresponding to the $u$-th UAV user is obtained by
\begin{equation}\label{received signal_MUMIMO}
\begin{split}
\hat{x}_u=\underbrace{\mathbf{g}_u^{\mathrm{H}} \mathbf{h}_u x_u}_{\text {Expected Signal }}+\underbrace{\sum_{u' \neq u} \mathbf{g}_u^{\mathrm{H}} \mathbf{h}_{u'} x_{u'}}_{ \text {Interference }} +\underbrace{\mathbf{g}_u^{\mathrm{H}} \mathbf{n}_{b} }_{\text {Noise }}.
\end{split}
\end{equation}

Accordingly, the post-equalization SINR for the $u$-th UAV is calculated as
\begin{equation}\label{SINR_UAV}
\begin{split}
\mathrm{SINR}_u=\frac{\left|\mathbf{g}_u^{\mathrm{H}} \mathbf{h}_u\right|^2}{\sum_{u' \neq u}\left|\mathbf{g}_u^{\mathrm{H}} \mathbf{h}_{u'}\right|^2+\sigma^2\left\|\mathbf{g}_u \right\|^2}.
\end{split}
\end{equation}

Based on Shannon capacity theory, the average achievable rate for the $u$-th UAV over all OFDM symbols and subcarriers can be expressed as
\begin{equation}\label{Mean_Rate_UAV}
\begin{split}
R_u = \frac{1}{KS}\sum_{k=1}^{K} \sum_{s=1}^{S} \log _2\left(1+ \mathrm{SINR}_u[k][s] \left( \mathbf{W}_{u}, a_{u, b} \right) \right).
\end{split}
\end{equation}

\subsection{Problem Formulation}

In multi-UAV cooperative perception tasks, each UAV captures different aerial views using onboard cameras. Instead of transmitting raw images to the ground server, these images are sparsified via Top-K pixel selection to enable efficient wireless transmission. The sparsified images are then encoded at the ground server, where multi-UAV feature fusion is performed to support downstream perception tasks, including semantic segmentation and instance segmentation.
The overall objective is to maximize the perception utility while minimizing the transmission latency, thereby maintaining high task-level accuracy under stringent communication resource constraints.

To evaluate the utility of semantic segmentation, the commonly used Intersection-over-Union (IoU) \cite{wang2020v2vnet} metric is adopted to measure the pixel-level overlap between the predicted segmentation and the ground-truth labels. Specifically, for the pixel region of the vehicle class, the IoU is computed as
\begin{equation}\label{IoU}
\begin{split}
\operatorname{IoU}\left(\hat{\ell}^{\mathrm{veh}}, \ell^{\mathrm{veh}}\right)=\frac{1}{T} \sum_{t=0}^{F-1} \frac{\sum_{h, w} \hat{\ell}_t^{\mathrm{veh}} \cdot \ell_t^{\mathrm{veh}}}{\sum_{h, w} \hat{\ell}_t^{\mathrm{veh}}+\ell_t^{\mathrm{veh}}-\hat{\ell}_t^{\mathrm{veh}} \cdot \ell_t^{\mathrm{veh}}},
\end{split}
\end{equation}
where $t\in\{0,\ldots,F-1\}$ denotes the frame index, and $F$ is the total number of frames in the perception and future prediction horizons.
$\hat{\ell}_t^{\mathrm{veh}}$ is the pixel of the predicted vehicle segmentation, and $\ell_t^{\mathrm{veh}}$ represents the pixel of the corresponding ground-truth label. $h$ and $w$ denote the height and width indices of the BEV image, respectively.

For the instance segmentation task, we adopt the panoptic quality (PQ) \cite{kirillov2019panoptic} metric to jointly evaluate segmentation accuracy and instance recognition performance, which is calculated by the product of segmentation quality (SQ) and recognition quality (RQ). Specifically, SQ measures the average IoU over matched prediction–ground-truth instance pairs $(p, g)$, while RQ reflects the detection accuracy of the instance matching. The PQ metric is computed as follows:
\begin{equation}\label{Recognition Quality}
\begin{split}
PQ = \underbrace{\frac{\sum_{(p, g) \in \mathrm{TP}} \text{IoU}(p, g)}{|{\mathrm{TP}}|}}_{\text{SQ}} \times \underbrace{\frac{|\mathrm{TP}|}{|\mathrm{TP}| + \frac{1}{2}|\mathrm{FP}| + \frac{1}{2}|\mathrm{FN}|}}_{\text{RQ}},
\end{split}
\end{equation}
where $\mathrm{TP}$, $\mathrm{FP}$, and $\mathrm{FN}$ represent the set of true positives (matched instances), false positives (unmatched predictions), and false negatives (missed ground truths), respectively. Consequently, the utility of PQ across the $F$ detection frames can be expressed as
\begin{equation}\label{Panoptic Quality 1}
\begin{split}
PQ = \sum_{t=1}^{F-1} \frac{\sum_{(p_t, g_t) \in {\mathrm{TP}}_{t}} \text{IoU}(p_t, g_t)}{|\mathrm{TP}_t| + \frac{1}{2}|\mathrm{FP}_t| + \frac{1}{2}|\mathrm{FN}_t|}.
\end{split}
\end{equation}

Subsequently, to reduce the transmission latency and communication overhead of UAV image data, the $\text{Top-K}$ strategy that transmits only a fraction $\kappa_{u}$ of the most informative pixels is adopted. Accordingly, the transmission data size $D_{u}$ for UAV 
$u$ can be calculated as
\begin{equation}\label{transmission data}
\begin{split}
D_{u} = \text{Top\_K}(u) = \kappa_{u}\cdot XY \cdot C \cdot M,
\end{split}
\end{equation}
where $X \times Y$ denotes the image resolution, $C$ is the number of channels, and $M$ represents the quantization bits per pixel. Based on the achievable rate given in \eqref{Mean_Rate_UAV}, the communication latency of UAV $u$ at frame $t$ can be computed as
\begin{equation}\label{transmission data1}
\begin{split}
L_u(t) = \frac{D_{u}(t)}{R_u(t)}.
\end{split}
\end{equation}

For the cooperative perception task, images from all UAVs are required to be received before joint processing. Consequently, we define the following objective function to balance the trade-off between weighted utility and maximum communication latency, expressed as
\begin{equation}\label{objective}
\begin{split}
\underbrace{ \alpha \text{U}_{\text{PQ}}(t) + (1-\alpha)\text{U}_{\text{IoU}}(t) }_{Weighted~Utility} - \lambda \cdot \underbrace{ \max_{u \in \mathcal{U}} \left( a_{u, b} \frac{D_{u}(t)}{R_u(t)} \right) }_{Comm.~lantency~L},
\end{split}
\end{equation}
where $\alpha \in [0,1]$ is the weighting factor between the PQ-based and IoU-based utilities, and $\lambda > 0$ is the penalty coefficient of the maximum latency. For the total $F$ frames, we formulate the multi-UAV cooperative perception problem as
\begin{subequations}\label{objective1}
\begin{align}
\label{o1} \max_{ \mathbf{W}_{u}, \kappa_{u}, a_{u,b} }&~ \frac{1}{F} \sum_{t=1}^{F}  
\alpha \text{U}_{\text{PQ}}(t) + (1-\alpha)\text{U}_{\text{IoU}}(t) - \lambda L(t) ~~~ \tag*{}\\
\label{o1a}\text { s.t. }~~~ &  1 \leq \text{Top-K}(u)  \leq XY \cdot C \cdot M, \quad \forall u \in \mathcal{U}, \\
\label{o1b}&  \mathbf{W}_{u} \in \mathcal{P}_{t_x}, \quad \forall u \in \mathcal{U}, \\
\label{o1c}&  \left\|\mathbf{W}_{u}\right\|_F=1,  \quad  \forall u \in \mathcal{U}, \\  
\label{o1d}&  a_{u, b} \in\{0, 1\}, \quad \forall u \in \mathcal{U}, 
\end{align}\notag
\end{subequations}
where constraint \eqref{o1a} limits the data size of transmitted UAV images after $\text{Top-K}$ selection. Constraint \eqref{o1b} selects $\mathbf{W}_{u}$ from a predefined precoder set $\mathcal{P}_{t_x}$ designed for single-stream data transmission with $t_x$ antennas and Constraint \eqref{o1c} normalizes $\mathbf{W}_{u}$ into a unit-power matrix, where $||\cdot||_F$ denotes the Frobenius norm. Constraint \eqref{o1d} indicates the UAV-BS association decisions for implementing the multi-UAV perception task. The optimization problem in \eqref{objective1} jointly determines the Top-K ratios, precoding matrices, and association indicators to maximize the weighted perception utility while minimizing the maximum communication latency.

\section{Perception Framework}

\begin{figure*}
\centering
\includegraphics[width=\textwidth]{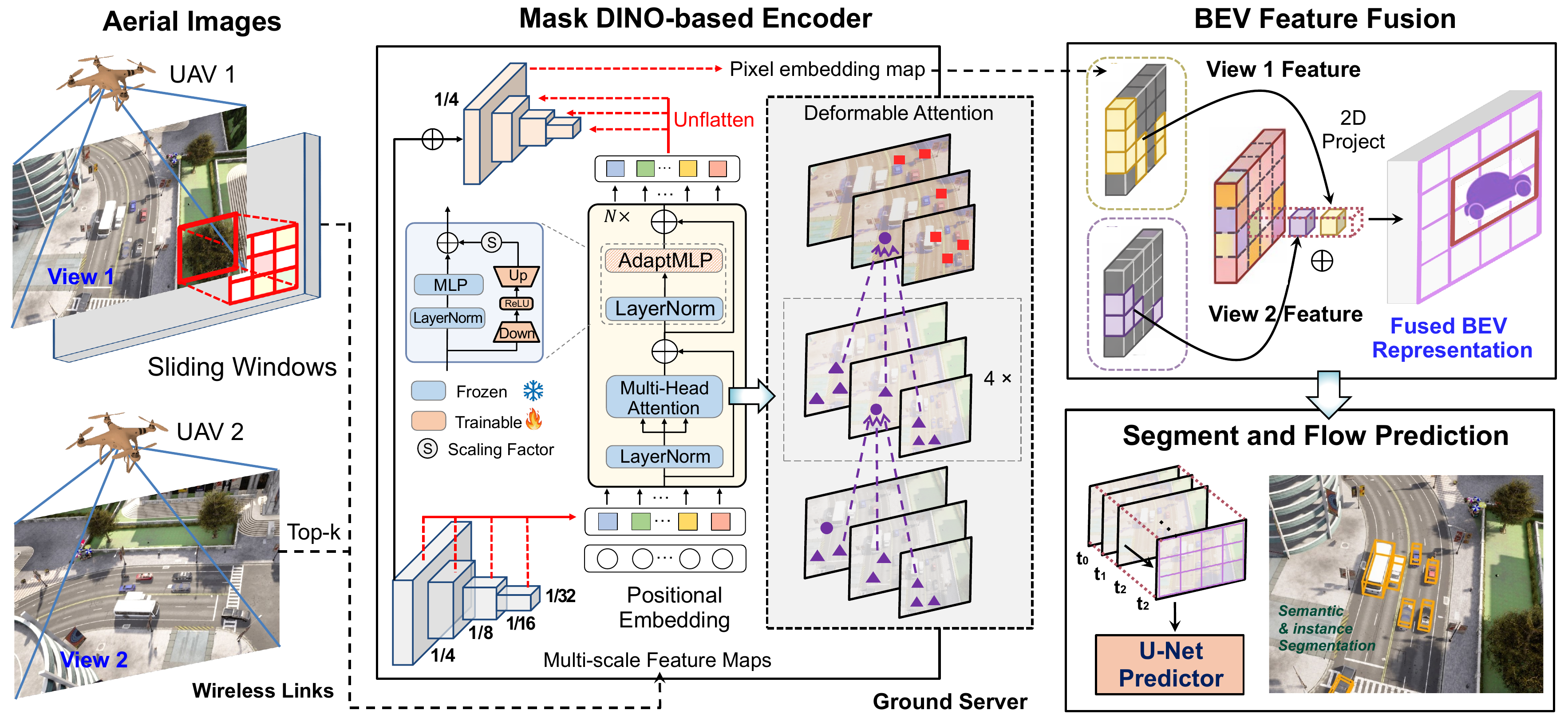} 
\centering
\captionsetup{justification=justified,format=plain}
\caption{ Illustration of the proposed BHU framework for multi-UAV cooperative perception. The aerial images are first sparsified via a Top-K selection mechanism and transmitted to the ground server through wireless links. A MaskDINO-based encoder extracts multi-scale features, which are projected into BEV representations and cooperatively fused across multi-UAVs. The fused BEV features are employed for downstream perception tasks, including vehicle instance segmentation and motion flow prediction.
}
\label{perception_framework}
\end{figure*}


This section presents the proposed BHU framework for multi-UAV cooperative perception. As illustrated in Fig.~\ref{perception_framework}, the framework consists of (i) a sliding-window Top-K selection scheme, (ii) a MaskDINO-based feature extractor, (iii) a BEV feature fusion module, and (iv) a U-Net predictor that outputs semantic and instance segmentation results. 

\subsection{Top-K Selection Scheme}

For an RGB image $\mathbf{I}_u \in \mathbb{R}^{C \times X \times Y}$ captured by UAV $u$, a lightweight neural network deployed onboard is employed to compress the image into a single-channel importance map, given by $\mathbf{S}_u = \phi_{RGB}(\mathbf{I}_u) \in \mathbb{R}^{1 \times X \times Y}$. Subsequently, for each pixel $\mathbf{I}_u(i,j)$ in the original image, a $(3 \times 3)$ sliding window, implemented as an averaging kernel with the center excluded, is applied to compute the average importance score of its eight neighboring pixels \cite{hu2022where2comm}. The score for $(i,j)$ is calculated  as follows:
\begin{equation}\label{importance score}
\begin{split}
\text{Score}_u(i, j)=\frac{1}{\left|\mathcal{N}_{i, j}\right|} \sum_{(i', j') \in \mathcal{N}_{i, j}} \mathbf{S}_u(i', j')-\mathbf{S}_u(i, j),
\end{split}
\end{equation}
where $\mathcal{N}_{i, j}$ denotes the set of scores within the $3\times3$ neighborhood centered at $(i, j)$. Given a Top-K ratio $\kappa_u \in (0,1]$ of UAV $u$,  the $\lfloor \kappa_u XY \rfloor $ pixels with the largest importance scores are selected for transmission, which can be expressed as
\begin{equation}\label{top_k score}
\begin{split}
\tilde{\mathbf{I}}_u(i,j) = \mathbb{I} \left( \text{Score}_u(i, j) \geq \text{S}_u^{\text{Top-K}} \right) \odot \mathbf{I}_u(i,j),
\end{split}
\end{equation}
where $\text{S}_u^{\text{Top-K}}$ denotes the$\lfloor \kappa_u XY \rfloor $-th largest importance score, $\mathbb{I}(\cdot) $ is the indicator function, and $\odot$ denotes the element-wise product operation. The sparsified image $\tilde{\mathbf{I}}_u(i,j)$ is then transmitted to a ground server via wireless links and is reconstructed into a complete image through a learnable  Gaussian interpolation module, which is computed by 
\begin{equation}\label{Gaussian interpolation}
\begin{split}
\hat{\mathbf{I}}_u(i,j)=\frac{ \sum_{(i', j') \in \mathcal{N}_{i, j}} \exp \left(-\frac{\|(i,j) -(i',j')\|_2^2}{2 \sigma_u^2}\right) \tilde{\mathbf{I}}_u(i’,j') }{ 
\sum_{(i', j') \in \mathcal{N}_{i, j} } \exp \left(-\frac{\|(i,j) -(i',j')\|_2^2}{2 \sigma_u^2}\right) }, 
\end{split}
\end{equation}
where $\hat{\mathbf{I}}_u \in \mathbb{R}^{C \times X \times Y}$ and $\sigma_u^2$ is a distance-weighting factor. 

\subsection{MaskDINO-based Feature Extractor}
Given the reconstructed image $\hat{\mathbf{I}}_u$, a MaskDINO-based visual encoder is employed to extract pixel-wise embeddings and multi-scale features for subsequent BEV projection. Specifically, a Swin-large Transformer \cite{liu2021swin} backbone first produces a pyramid of multi-scale feature maps as
\begin{equation}\label{multi-scale feature maps}
\begin{split}
\left\{\mathbf{Z}_u^{(l)}\right\}_{l=1}^L=\phi_{\text{Swin-L}}\left(\hat{\mathbf{I}}_u\right), \quad \mathbf{Z}_u^{(l)} \in \mathbb{R}^{C_l \times \frac{X}{d_l} \times \frac{Y}{d_l}},
\end{split}
\end{equation}
where $L$ denotes the number of pyramid levels, and $d_l$ is the downsampling factor at level $l$. Subsequently, a deformable attention module is adopted to efficiently aggregate multi-scale features by sampling a sparse set of reference points $\mathbf{p}$ across feature levels, expressed as follows: 
\begin{equation}\label{deformable attention}
\begin{split}
\widehat{\mathbf{Z}}_u=\text { DeformAttn }\left(\left\{\mathbf{Z}_u^{(l)}\right\}_{l=1}^L,  \mathbf{p} \right) \text {, }
\end{split}
\end{equation}
where the $\text{DeformAttn}(\cdot)$ for the $q$-th query $\mathbf{z}_q$ of $\{\mathbf{Z}_u^{(l)}\}_{l=1}^L$  can be calculated by 
\begin{equation}\label{deformable attention 1}
\begin{split}
\text{DeformAttn}\left(\mathbf{z}_q, \mathbf{p}_q\right)=\sum_{l=1}^L \sum_{n=1}^{N_{\text{ref} }} \mathbb{A}_{q, l, n} \phi_v \mathbf{Z}_u^{(l)}\left(\mathbf{p}_q+\Delta \mathbf{p}_{q, l, n}\right),
\end{split}
\end{equation}
where $N_{\text{ref}}$ denotes the number of sampling points per level, $\mathbb{A}_{q, l, n}$ is the attention weight that satisfies $\sum_{l,n}\mathbb{A}_{q, l, n} =1$. $\Delta \mathbf{p}_{q, l, n}$ is the learned offset, and $\phi_v$ is a linear projection applied to the sampled feature. Finally, the output feature maps in $\widehat{\mathbf{Z}}_u$ are unflatten and concatenated into a dense pixel embedding map $\widehat{\mathbf{Z}}_u^e \in \mathbb{R}^{C_e \times X' \times Y'}$, serving as the input to the BEV projection and cross-UAV feature fusion module.

\subsection{BEV Feature Fusion}

Following the Lift-Splat-Shoot (LSS) \cite{chen2025semantic} paradigm, the pixels $(i, j) \in \{0, \ldots, X'-1 \} \times \{0, \ldots, Y'-1 \}$  
in the embedding map $\widehat{\mathbf{Z}}_u^e$ are projected into a unified LiDAR coordinate system where the BEV features across UAVs are fused. 
Given the camera intrinsic matrix, 
\begin{equation}\label{intrinsic}
\begin{split}
\mathbb{K} = \begin{bmatrix} f_x & 0 & i_c \\ 0 & f_y & j_c \\ 0 & 0 & 1 \end{bmatrix} ~ \in \mathbb{R}^{3 \times 3},
\end{split}
\end{equation}
where $f_x$ and $f_y$ denote the focal length and $i_c, j_c$ represents the principal point of the feature map,$(i,j)$ can be transformed into a corresponding 3D ray $[X_c/Z_c, Y_c/Z_c, 1]^T$ in the camera coordinate system as follows:
\begin{equation}\label{3D ray}
\begin{split}
\underbrace{
\begin{bmatrix}
\frac{1}{f_x} & 0 & -\frac{i_c}{f_x} \\
0 & \frac{1}{f_y} & -\frac{j_c}{f_y} \\
0 & 0 & 1
\end{bmatrix}
}_{\mathbb{K}^{-1}}
\cdot
\begin{bmatrix}
i \\
j \\
1
\end{bmatrix}
=
\begin{bmatrix}
\frac{i - i_c}{f_x} \\
\frac{j - j_c}{f_y} \\
1
\end{bmatrix}
=
\begin{bmatrix}
X_c / Z_c \\
Y_c / Z_c \\
1
\end{bmatrix},
\end{split}
\end{equation}
where $[X_c, Y_c, Z_c]^T$ represents the 3D camera coordinates.  



Let the extrinsic matrix from the camera coordinate to the LiDAR coordinate be $\mathbb{E} \in \mathbb{R}^{4 \times 4}$, the corresponding 3D point in the LiDAR coordinate system is given by
\begin{equation}\label{extrinsic}
\begin{split}
\begin{bmatrix}
X_{\text{lidar}} \\
Y_{\text{lidar}} \\
Z_{\text{lidar}} \\
1
\end{bmatrix}
=
\underbrace{
\begin{bmatrix} \mathbf{R_E} & \mathbf{T_E} \\  \\ \mathbf{0}^T & 1 \end{bmatrix}
}_{ \mathbb{E} \in \mathbb{R}^{4 \times 4}}
\cdot
\begin{bmatrix}
X_c  \\
Y_c  \\
 Z_c \\
1
\end{bmatrix},
\end{split}
\end{equation}
where $\mathbf{R_E} \in \mathbb{R}^{3 \times 3} $ and $\mathbf{T_E} \in \mathbb{R}^{3 \times 1}$ denote the rotation matrix and the translation vector, respectively. 

Defining a 2D BEV plane on the LiDAR coordinate system with a predefined spatial range $[h_{\min}, h_{\max}] \times [w_{\min}, w_{\max}]$ and resolution $\Delta_h, \Delta_w$, we can obtain the BEV length and width as follows:
\begin{equation}\label{BEV grid}
\begin{split}
W_{\text{bev}} = \left\lfloor \frac{w_{\max} - w_{\min}}{\Delta_w} \right\rfloor , \quad H_{\text{bev}} = \left\lfloor \frac{h_{\max} - h_{\min}}{\Delta_h} \right\rfloor, 
\end{split}
\end{equation}

Accordingly, each 3D point $(X_{\text{lidar}}, Y_{\text{lidar}}, Z_{\text{lidar}} )$ can be assigned and projected to a 2D BEV grid $(w, h)$ by
\begin{equation}\label{BEV project}
\begin{split}
w = \left\lfloor \frac{X_{\text{lidar}} - w_{\min}}{\Delta_w} \right\rfloor , \quad h = \left\lfloor \frac{Y_{\text{lidar}} - h_{\min}}{\Delta_h} \right\rfloor. 
\end{split}
\end{equation}

Consequently, following the geometric relationship defined by the camera intrinsic and extrinsic, the pixel embedding map $\widehat{\mathbf{Z}}_u^e \in \mathbb{R}^{C_e \times X' \times Y'}$ of the MaskDINO encoder are transformed into a BEV domain representation $\mathbf{BEV}_u \in \mathbb{R}^{C_e \times W_{\text{bev}} \times H_{\text{bev}}}$. For the perception frame $t$, the BEV feature maps from all participating UAVs are aggregated by performing element-wise summation at each spatial location $(h,w)$, constructing a fused BEV representation as
$\mathbf{BEV}_t = \sum_{u=1}^{U}\mathbf{BEV}_u(t)$.

\subsection{Multi-Task Prediction}

To jointly perform the semantic segmentation and instance segmentation tasks in multi-frame perception,
two parallel prediction heads are appended to a shared U-Net-based decoder, which reconstructs segmentation and motion representations from the fused BEV feature maps $\{\mathbf{BEV}_t\}_{t=0}^{F}$. For semantic segmentation, the decoder produces pixel-level class predictions for vehicle and background regions in the BEV plane. The task is modeled as a pixel-wise classification problem \cite{wang2024drones} and optimized using the cross-entropy loss function:
\begin{equation}\label{cross-entropy}
\begin{split}
\mathcal{L}_{\mathrm{seg}} (\hat{\ell}_t^{\mathrm{veh}}, \ell_t^{\text {veh }} )  =  \frac{1}{2\sigma_{\mathrm{seg}}^2} \mathcal{L}_{CE} (\hat{\ell}_t^{\mathrm{veh}}, \ell_t^{\text {veh}} ) + \frac{1}{2} \log \sigma_{\mathrm{seg}}^2, 
\end{split}
\end{equation}
where $\log \sigma_{\mathrm{seg}}^2$ is a learnable uncertainty parameter that can adaptively adjust the loss weight for multi-task optimization.

For instance segmentation, let $\hat{\ell}_t^{\text {flow }}$ be the predicted instance flow in the BEV plane at frame $t$, and $\ell_t^{\text {flow }}$ be the corresponding instance ground truth. We adopt the $L_1$ regression with a similar learnable parameter $\log \sigma_{\mathrm{flow}}^2$ as the loss function given by 
\begin{equation}\label{L_1}
\begin{split}
\mathcal{L}_{l_1} (\hat{\ell}_t^{\text {flow }}, \mathbf{\ell}_t^{\text {flow }} ) = \frac{1}{2\sigma_{\mathrm{flow}}^2} \left\|\hat{\mathbf{\ell}}_t^{\text {flow }}-\mathbf{\ell}_t^{\text {flow }}\right\|_1  + \frac{1}{2} \log \sigma_{\mathrm{flow}}^2.
\end{split}
\end{equation}

For the total $F$ frames, the overall loss function is defined as the weighted summation of the semantic segmentation and instance flow losses over all frames as
\begin{equation}\label{Loss BEV}
\begin{split}
\mathcal{L}=\frac{1}{F} \sum_{t=0}^{F} \gamma^t\left(\lambda_{\mathrm{seg}} \mathcal{L}_{\mathrm{seg}} (\hat{\ell}_t^{\mathrm{veh}}, \ell_t^{\text {veh }} ) +\lambda_{\mathrm{flow}} \mathcal{L}_{l_1} (\hat{\ell}_t^{\text {flow }}, \ell_t^{\text {flow }} ) \right) ,
\end{split}
\end{equation}
where $\gamma^t$ is a temporal factor that discounts the contribution of long-horizon future frames, and $\lambda_{\mathrm{seg}}$ and  $\lambda_{\mathrm{flow}}$ are coefficients that balance the relative importance of semantic segmentation and instance flow regression.

\section{Proposed Solution}

After training the BHU framework, cooperative UAVs can perform real-time inference by transmitting Top-K sparsified images to the ground server, where the corresponding perception performance of IoU and PQ metrics can be evaluated for the objective in \eqref{objective1}. To solve the perception utility and latency trade-off problem formulated in \eqref{objective1}, this section proposes a DDIM-based DRL algorithm to jointly optimize the Top-K selection ratio, UAV association, and precoding strategy.

\subsection{Deep Reinforcement Learning Framework}
Owing to the stochastic wireless channel dynamics during UAV image transmission and the involvement of binary UAV-BS association variables in the numerator and denominator of the SINR expressions, the problem \eqref{objective1} constitutes a stochastic mixed-integer programming problem. Such an optimization problem has been demonstrated to be non-convex and NP-hard, making it difficult to obtain an optimal solution \cite{yunting2}. Therefore, we adopt a DRL framework to independently output the UAV-BS association, the top-K ratio, and the precoding schemes, thus providing a feasible solution for the formulated problem in \eqref{objective1}. The state, action, and reward function of the DRL framework are defined as follows.

\subsubsection{State}
In the multi-UAV cooperative perception scenario, the trade-off between perception utility and transmission latency depends on both the wireless channel states and the images captured by the UAVs at the current frame $t$. Accordingly, the wireless channels $\mathbf{H}_{u,b}$ between all UAVs and the BS, together with the images $\mathbf{I}_u$ captured by each UAV, are incorporated as the state information of the DRL framework, expressed as
\begin{equation}\label{state}
\begin{split}
\mathcal{S}_t=\left(\left\{\mathbf{H}_{u, b}(t)\right\}_{u=1}^U,\left\{\mathbf{I}_u(t)\right\}_{u=1}^U\right).
\end{split}
\end{equation}

\subsubsection{Action}

The action to solve the problem \eqref{objective1} includes the selection of cooperative UAVs $a_{u,b}$, Top-K ratios $\kappa_{u}$, and the precoding matrix $\mathbf{W}_{u}$ for each UAV. Denote the DRL action associated with the UAV selection as $\mathcal{A}_{\mathcal{U}}$, which can be obtained using the binomial coefficient $\binom{U}{N_\text{uav}}$ that selects $N_\text{uav}$ UAVs from a total of $U$ UAVs. Consequently, by summing all the binomial coefficients with ${N_\text{uav}} = \{1, \ldots, U\}$, the action space corresponding to the UAV-BS association scheme can be expressed as
\begin{equation}\label{UAV selection}
\begin{split}
| \mathcal{A}_{\mathcal{U}}| = \sum_{N_\text{uav} = 1}^{U} \binom{U}{N_\text{uav}} = \sum_{N_\text{uav} = 1}^{U} \frac{U!}{N_\text{uav}!(U-N_\text{uav})!}.
\end{split}
\end{equation}

For the action of selecting the Top-K transmission ratio, we discretize the ratio into a finite set $\mathbb{K}=\{\kappa_{\text{min}} +(n-1) \Delta_{\kappa} \}_{n=1}^{N_{\kappa}}$ that consists of $N_{\kappa}$ candidate values, where $\Delta_{\kappa}$ denotes the interval between adjacent ratios. Each ratio in $\mathbb{K}$ lies within the range $[0,1]$ to satisfy constraint \eqref{o1a}. Accordingly, the action $\mathcal{A}_{\kappa}$ for the Top-K ratio selection is defined as follows:
\begin{equation}\label{TopK selection}
\begin{split}
\mathcal{A}_{\kappa}=\left\{ \kappa_u \in \mathbb{K}, ~ \forall u \in\{1, \ldots, U\}\right\}.
\end{split}
\end{equation}

With the total number of $U$ UAVs, the action space for the Top-K ratio selection is given by $| \mathcal{A}_{\kappa}| = N_{\kappa}^U$.

For the action of precoding matrix selection $\mathcal{A}_{\mathcal{W}}$, the predefined precoder set $\mathcal{P}_{t_x}$ in constraint \eqref{o1b} is constructed based on the Type-I codebook specified in the 3GPP standard \cite{3GPP38.214}. Each precoding matrix in the Type-I codebook is composed of discrete Fourier transform (DFT) beamforming vectors with unit power, thereby satisfying the unit norm constraint in \eqref{o1c}. The composition of Type-I codebook is provided as follows.

\noindent\textbf{Remark 1:} 
For a uniform planar array with $(N_x \times N_y)$ antenna elements, let $O_{x}$  and $O_{y}$ denote the oversampling factors in the horizontal and vertical dimension, the corresponding discrete DFT beamforming vectors can be expressed as 
\begin{equation}\label{B1}
\begin{split}
\mathcal{C}_{n_x} =& \left[1 ~e^{ j 2 \pi \frac{ n_x  \cdot 1}{N_{x} O_{x}}} \cdots ~ e^{j 2 \pi \frac{ n_x \cdot (N_{x}-1)}{N_{x} O_{x}}}\right]^{\mathrm{T}}, \\
& \quad\quad \forall n_x \in \{0,\cdots,O_{x}N_{x}-1\},
\end{split}
\end{equation}
\begin{equation}\label{B2}
\begin{split}
\mathcal{C}_{n_y} =& \left[1 ~e^{ j 2 \pi \frac{ n_y  \cdot 1}{N_{y} O_{y}}} \cdots ~ e^{j 2 \pi \frac{ n_y \cdot (N_{y}-1)}{N_{y} O_{y}}}\right]^{\mathrm{T}}, \\
& \quad\quad \forall n_y \in \{0, \cdots,O_{y}N_{y}-1\}.
\end{split}
\end{equation}

Subsequently, the precoding matrix for UAV $u$ is calculated by 
\begin{equation}\label{B2}
\begin{split}
\mathbf{W}_{u} = 
 \left[\begin{array}{cc}
\mathcal{C}_{n_x} \otimes \mathcal{C}_{n_y} & 0 \\
0 & \mathcal{C}_{n_x} \otimes \mathcal{C}_{n_y}
\end{array}\right]
\mathbf{W}_{\text{adj}},
\end{split}
\end{equation}
where $\mathbf{W}_{\text{adj}}$ is an adjustment matrix with narrow-band properties. For the precoding matrix designed for single-stream transmission, $\mathbf{W}_{\text{adj}}$ is formulated as
\begin{equation}\label{B3}
\begin{split}
\mathbf{W}_{\text{adj}} = \frac{1}{\sqrt{2N_{x}N_{y}}} 
 \left[\begin{array}{c}
\mathbf{e} \\
\psi_{m} \mathbf{e}
\end{array}\right]^\mathrm{T}.
\end{split}
\end{equation} 

Specifically, $\psi_{m}$ is a co-phasing factor and is obtained by
\begin{equation}\label{B4}
\begin{split}
\psi_{m} &= e^{j \pi \frac{m}{2}}, \quad m = \{0,~1,~2,~3\},
\end{split}
\end{equation}
where the values of $m$ index four discrete co-phasing candidates $\psi_{m} \in \{1,j,-1,-j\}$.

Let $\mathcal{M}_{n_x, n_y}$ be the Kronecker product of the DFT vectors:
\begin{equation}\label{B5}
\begin{split}
\mathcal{M}_{n_x, n_y} = \mathcal{C}_{n_x} \otimes \mathcal{C}_{n_y} \in \mathbb{C}^{N_xN_y \times 1}, %
\end{split}
\end{equation}
the candidate precoding matrix in $\mathcal{P}_{t_x}$ can be expressed as
\begin{align}\label{B6}
\mathbf{W}_u(n_x, n_y, m) = \frac{1}{\sqrt{2N_{x}N_{y}}}\left[\begin{array}{c}
\mathcal{M}_{n_x, n_y} \\
\psi_{n_3} \mathcal{M}_{n_x, n_y}
\end{array}\right].
\end{align}

Therefore, for the total number of $U$ UAVs, the action space for the precoding selection is calculated by $| \mathcal{A}_{\mathcal{W}}| = |\mathcal{P}_{t_x}|^U$.

\subsubsection{Reward function}
Given the state $\mathcal{S}_t$ at frame $t$, after selecting the cooperative UAVs, Top-K ratios, and precoding matrices, the reward is calculated as the difference between weighted utility and maximum transmission latency, given by
\begin{equation}\label{reward}
\begin{split}
& \mathcal{R} (\mathcal{S}_t, \mathcal{A}^t_{\mathcal{U}}, \mathcal{A}_{\kappa}^t, \mathcal{A}^t_{\mathcal{W}} ) = \\
&~~~\alpha \text{U}_{\text{PQ}}(t) + (1-\alpha)\text{U}_{\text{IoU}}(t)  - \lambda \cdot \max_{u \in \mathcal{U}} \left( a_{u, b} \frac{D_{u}(t)}{R_u(t)} \right).
\end{split}
\end{equation}

Consequently, maximizing the reward function is equivalent to optimizing the objective in problem \eqref{objective1}, thereby enhancing the performance of perception tasks while reducing the UAV image transmission latency.

\begin{figure}
\centering
\includegraphics[width=0.49\textwidth]{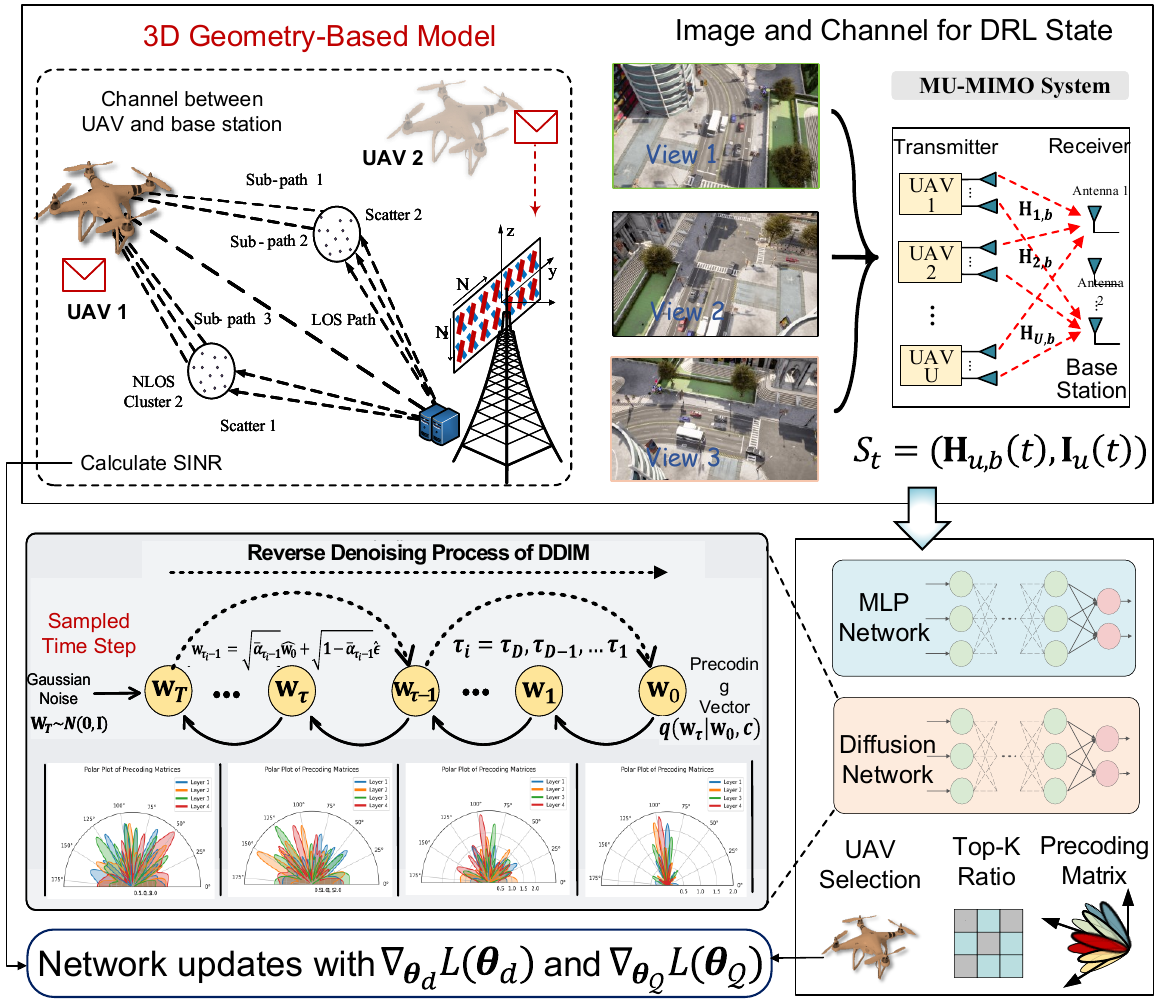} 
\centering
\captionsetup{justification=justified,format=plain}
\caption{ Illustration of the proposed DDIM-based DRL framework. An MLP network is employed to determine the cooperative UAVs and Top-K sparsification ratios, while a DDIM module is employed to generate precoding actions by modeling the conditional distribution of optimal precoding vectors through a reverse denoising process. }
\label{DDIM_DRL}
\end{figure}

\subsection{DDIM-based Algorithm}
Once the number of cooperative UAVs is determined, the action spaces associated with the Top-K ratio  $(| \mathcal{A}_{\kappa}| = N_{\kappa}^U)$ and the precoding matrix $(|\mathcal{A}_{\mathcal{W}}|=|\mathcal{P}_{t_x}|^U)$ grow exponentially with the number of UAVs. As a result, directly allowing the DRL to output a policy over the joint action space would incur prohibitive dimensionality and computational complexity. To address this issue, we first employ a neural network to determine the Top-K transmission ratio for each selected UAV. Based on the actions of selected UAVs $\mathcal{A}_{\mathcal{U}}$ and Top-K ratios $\mathcal{A}_{\kappa}$, we can exhaustively search the precoding codebook for each UAV to obtain the optimal precoding matrices $
\{\mathbf{W}_{u}^{o}\}_{u=1}^{U}$ for $\mathcal{A}_{\mathcal{W}}$. Subsequently, we employ a diffusion model to model the distribution of optimal precoding decisions, thereby enabling the generation of precoding actions.

The overall DRL framework is depicted in Fig.~\ref{DDIM_DRL}. Although diffusion models have demonstrated strong capabilities in decision modeling and generation, conventional diffusion-based approaches typically require a large number of iterative denoising steps, which significantly increases training overhead and degrades training efficiency. Consequently, we incorporate the DDIM mechanism \cite{song2020denoising}, which approximates the denoising process of diffusion models by sampling a subsequence of time steps, thereby substantially improving the training efficiency of the proposed DRL framework.

Let $\mathbf{w}_0$ denote the flattened and concatenated vector of the optimal precoding matrices $
\{\mathbf{W}_{u}^{o}\}_{u=1}^{U}$. The function of DDIM is to model the conditional distribution of $\mathbf{w}_0$ given the UAV selection actions $\mathcal{A}_{\mathcal{U}}$ and the sparsification ratios $\mathcal{A}_{\kappa}$, and then to generate the corresponding precoding vectors $\bar{\mathbf{w}}_0$ for action $\mathcal{A}_{\mathcal{W}}$. 
In the forward noising process of DDIM, following the same procedure as the DDPM, Gaussian noise is progressively injected into the $\mathbf{w}_0$ vector over $T$ time steps, where the noise added at time step $\tau$ has fixed variance $\beta_{\tau}$. For the denoising process, a subsequence $\{ \tau_i \}_{i=1}^{\mathcal{D}}$ is sampled from the original diffusion timeline of length $T$, where $ \mathcal{D} \ll T$ denotes the total number of DDIM denoising steps. 
Consequently, let $\mathbf{c}$ be the flattened and concatenated conditional vector of $\mathcal{A}_{\mathcal{U}}$ and $\mathcal{A}_{\kappa}$, the forward noising adding process of DDIM is described as
\begin{equation}\label{forward process}
\begin{split}
\boldsymbol{q} \left( \mathbf{w}_{\tau} | \mathbf{w}_{\tau - 1}, \mathbf{c} \right) = \mathcal{N}\left( \mathbf{w}_{\tau} ; \sqrt{1-\beta_{\tau}} \mathbf{w}_{\tau-1},\beta_{\tau} \boldsymbol{I}\right).
\end{split}
\end{equation}

Subsequently, the distribution of the $\tau$-th noised precoding vector is calculated as 
\begin{equation}\label{noised precoding}
\begin{split}
\boldsymbol{q} \left(\mathbf{w}_{\tau} | \mathbf{w}_0, \mathbf{c} \right) 
& =\prod_{t = 1}^{\tau}  \mathcal{N}\left(\mathbf{w}_{\tau} ; \sqrt{1-\beta_{\tau}} \mathbf{w}_{\tau-1}, \beta_{\tau} \boldsymbol{I}\right) \\
& \stackrel{(1)}{=} \mathcal{N}\left(\mathbf{w}_{\tau} ; \sqrt{\bar{\alpha}_{\tau}} \mathbf{w}_{0}, \left( 1-\bar{\alpha}_{\tau} \right) \boldsymbol{I}\right),
\end{split}
\end{equation}
where $\stackrel{(1)}{=}$ is obtained by letting $\alpha_{\tau} = 1-\beta_{\tau}$ and $ \bar{\alpha}_{\tau} = \prod_{1}^{\tau} \alpha_{\tau}$. As a result, for an arbitrary time step $\tau$, the noised precoding vector $\mathbf{w}_{\tau}$ can be directly computed as
\begin{equation}\label{noised precoding}
\begin{split}
\mathbf{w}_{\tau} = \sqrt{\bar{\alpha}_{\tau}} \mathbf{w}_{0} + \sqrt{\left( 1-\bar{\alpha}_{\tau} \right)} \boldsymbol{\epsilon}_{\tau}, \quad \boldsymbol{\epsilon}_{\tau} \sim \mathcal{N}\left(\mathbf{0}, \boldsymbol{I}\right).
\end{split}
\end{equation}

In the denoising process of the diffusion model, given the original precoding vector $\mathbf{w}_{0}$ and the conditioning variable $\mathbf{c}$, the posterior distribution of the previous time step $\tau-1$ can be derived from Bayes' rule as
\begin{equation}\label{intermediate reverse transition}
\begin{split}
\boldsymbol{q}\left(\mathbf{w}_{\tau-1} | \mathbf{w}_{\tau}, \mathbf{w}_{0}, \mathbf{c} \right) &= \boldsymbol{q} (\mathbf{w}_{\tau} | \mathbf{w}_{\tau-1}, \mathbf{c} ) \frac{\boldsymbol{q} (\mathbf{w}_{\tau-1} | \mathbf{w}_{0}, \mathbf{c} )}{\boldsymbol{q} (\mathbf{w}_{\tau} | \mathbf{w}_{0}, \mathbf{c} )} \\
=& \mathcal{N}\left(\mathbf{w}_{\tau-1} ; \boldsymbol{\mu}_{\tau}\left(\mathbf{w}_{\tau}, \mathbf{w}_{0}, \mathbf{c} \right), \boldsymbol{\sigma}_{\tau}^2 \boldsymbol{I}\right),
\end{split}
\end{equation}
where the mean $\boldsymbol{\mu}_{\tau}$ and the variance $\boldsymbol{\sigma}_{\tau}^2$ are given by
\begin{equation}\label{mu}
\begin{split}
\boldsymbol{\mu}_{\tau} = \frac{\sqrt{\bar{\alpha}_{\tau-1}} \beta_{\tau}}{1-\bar{\alpha}_{\tau}} \mathbf{w}_0+\frac{\sqrt{\alpha_{\tau}}\left(1-\bar{\alpha}_{{\tau}-1}\right)}{1-\bar{\alpha}_{\tau}} \mathbf{w}_{\tau},
 \end{split}
\end{equation}
\begin{equation}\label{sigma}
\begin{split}
\boldsymbol{\sigma}_{\tau}^2 = \frac{1-\bar{\alpha}_{\tau-1}}{1-\bar{\alpha}_{\tau}} \beta_{\tau}.
\end{split}
\end{equation}

However, $\mathbf{w}_{0}$ is not available during the practical reverse process. Therefore, a neural network $\boldsymbol{\theta}_d$ is employed to estimate $\mathbf{w}_{0}$ with $\hat{\mathbf{w}}_{0}(\mathbf{w}_{\tau}, \tau, \mathbf{c} ; \boldsymbol{\theta}_d)$.  Accordingly, we can obtain a mean $\tilde{\boldsymbol{\mu}}_{\tau}(\boldsymbol{\theta}_d)$ by substituting $\hat{\mathbf{w}}_{0}$ into Eq.~\eqref{mu}, and the resulting posterior distribution can be written as
\begin{equation}\label{posterior distribution 1}
\begin{split}
\boldsymbol{p}_{\boldsymbol{\theta}_d}\left(\mathbf{w}_{\tau-1} | \mathbf{w}_{\tau}, \mathbf{c} \right)=\mathcal{N}\left(\mathbf{w}_{\tau-1} ; \tilde{\boldsymbol{\mu}}_{\tau}\left(\boldsymbol{\theta}_d \right), \boldsymbol{\sigma}_{\tau}^2 \boldsymbol{I}\right).
\end{split}
\end{equation}

Subsequently, $\boldsymbol{\theta}_d$ is optimized by minimizing the Kullback-Leibler (KL) divergence between $\boldsymbol{q}\left(\mathbf{w}_{\tau-1} | \mathbf{w}_{\tau}, \mathbf{w}_{0}, \mathbf{c} \right)$ and $\boldsymbol{p}_{\boldsymbol{\theta}_d}\left(\mathbf{w}_{\tau-1} | \mathbf{w}_{\tau}, \mathbf{c} \right)$, i.e.,
\begin{equation}\label{reverse KL divergence}
\begin{split}
 \underset{ \boldsymbol{\theta}_d }{\arg \min } & ~ D_{\mathrm{KL}}\left( \boldsymbol{q}\left(\mathbf{w}_{\tau-1} | \mathbf{w}_{\tau}, \mathbf{w}_{0}, \mathbf{c} \right) \| \boldsymbol{p}_{\boldsymbol{\theta}_d}\left(\mathbf{w}_{\tau-1} | \mathbf{w}_{\tau}, \mathbf{c} \right) \right) \\
= & \mathcal{N}\left(\mathbf{w}_{\tau-1} ; \boldsymbol{\mu}_{\tau}, \boldsymbol{\sigma}_{\tau}^2 \boldsymbol{I}\right) \| \mathcal{N}\left(\mathbf{w}_{\tau-1} ; \tilde{\boldsymbol{\mu}}_{\tau}\left(\boldsymbol{\theta}_d \right), \boldsymbol{\sigma}_{\tau}^2 \boldsymbol{I}\right)  \\
= & \frac{1}{2}\left(\left( \tilde{\boldsymbol{\mu}}_{\tau}\left(\boldsymbol{\theta}_d \right) - \boldsymbol{\mu}_{\tau} \right)^T (\boldsymbol{\sigma}_{\tau}^2)^{-1} \left( \tilde{\boldsymbol{\mu}}_{\tau}\left(\boldsymbol{\theta}_d \right) - \boldsymbol{\mu}_{\tau} \right)\right) \\
= & \frac{1}{2 \tilde{\boldsymbol{\sigma}}_{\tau}^2 } \left\| \tilde{\boldsymbol{\mu}}_{\tau}\left(\boldsymbol{\theta}_d \right) - \boldsymbol{\mu}_{\tau}  \right\|^2,\\
= & \frac{1}{2 \tilde{\boldsymbol{\sigma}}_{\tau}^2}  \frac{\bar{\alpha}_{\tau-1} \beta_{\tau}^2}{\left(1-\bar{\alpha}_{\tau}\right)^2} \left\| \hat{\mathbf{w}}_{0}(\mathbf{w}_{\tau}, \tau, \mathbf{c}) -\mathbf{w}_{0} \right\|^2.
\end{split}
\end{equation}

In this paper, we use the following simplified expression as the training loss function of the diffusion model:
\begin{equation}\label{Difussion loss}
\begin{split}
\mathcal{L}(\boldsymbol{\theta}_d ) = \nabla_{\boldsymbol{\theta}_d} \mathbb{E}_{\mathbf{w}_{\tau}, \tau, \mathbf{c} }\left[\left\| \hat{\mathbf{w}}_{0}(\mathbf{w}_{\tau}, \tau, \mathbf{c}) -\mathbf{w}_{0} \right\|^2\right].
\end{split}
\end{equation}

For the DDIM denoising process, we denote the sampled time-step sequence as
\begin{equation}\label{time series}
\begin{split}
\{ \tau_i \}_{i=1}^{\mathcal{D}} = \{ \tau_D > \tau_{D-1} > \cdots > \tau_1 \}.
\end{split}
\end{equation}

\begin{algorithm}[t]
    \caption{DDIM-based DRL Algorithm. }\label{algorithm1}
$\textbf{Input:}$ The channel matrix and RGB images.\\
$\textbf{Ouput:}$ The UAV-BS association, Top-K ratios, and precoding scheme.\\
\Repeat{ The networks $\boldsymbol{\theta}_{d}$ and $\boldsymbol{\theta}_{\mathcal{Q}}$ converges}
{ Input the state $\mathcal{S}_t= (\left\{\mathbf{H}_{u, b}(t)\right\}_{u=1}^U,\left\{\mathbf{I}_u(t)\right\}_{u=1}^U )$; \\
  Select an action for UAV association $\mathcal{A}^i_{\mathcal{U}}$ and Top-K ratios $\mathcal{A}_{\kappa}$ based on the output $\mathcal{Q}$-values; \\
  Exhaustively search the codebook $\mathcal{P}_{t_x}|$ to obtain the precoding vector $\mathbf{w}_{0}$; \\
  Sample $\tau \sim \operatorname{Uniform}(\{1, \ldots, T\})$; \\
  Calculate the noised the precoding vector $\mathbf{w}_{\tau} = \sqrt{\bar{\alpha}_{\tau}} \mathbf{w}_{0} + \sqrt{\left( 1-\bar{\alpha}_{\tau} \right)} \boldsymbol{\epsilon}_{\tau}, ~ \boldsymbol{\epsilon}_{\tau} \sim \mathcal{N}\left(\mathbf{0}, \boldsymbol{I}\right)$; \\
  Update diffusion network $\boldsymbol{\theta}_d$ based on $\nabla_{\boldsymbol{\theta}_d} \mathbb{E}_{\mathbf{w}_{\tau}, \tau, \mathbf{c} }\left[\left\| \hat{\mathbf{w}}_{0}(\mathbf{w}_{\tau}, \tau, \mathbf{c}) -\mathbf{w}_{0} \right\|^2\right]$;\\
\For{DDIM denoising step $\tau_i = \tau_D, \tau_{D-1},\cdots,1$ }
{ Obtain $ \mathbf{w}_{\tau_{i-1}}=\sqrt{\bar{\alpha}_{\tau_{i-1}}} \hat{\mathbf{w}}_0(\mathbf{w}_{\tau_i}, \tau_i, \mathbf{c}) + \sqrt{1-\bar{\alpha}_{\tau_{i-1}} } \hat{\boldsymbol{\epsilon}}$;
}
   Generate $\bar{\mathbf{w}}_0$ for precoding action $\mathcal{A}_{\mathcal{W}}$; \\

  Obtain reward $\mathcal{R} (\mathcal{S}_t, \mathcal{A}^t_{\mathcal{U}}, \mathcal{A}_{\kappa}^t, \mathcal{A}^t_{\mathcal{W}} ) $ using Eq.~\eqref{reward}; \\
  Update network $\boldsymbol{\theta}_{\mathcal{Q}}$ based on $\mathcal{L}(\boldsymbol{\theta}_{\mathcal{Q}})
= \frac{1}{2}  \left\| \mathcal{Q}(\mathcal{S}^t,\mathcal{A}^t_{\mathcal{U}}, \mathcal{A}_{\kappa}^t)- \mathcal{R} (\mathcal{S}_t, \mathcal{A}^t_{\mathcal{U}}, \mathcal{A}_{\kappa}^t, \mathcal{A}^t_{\mathcal{W}} )  \right\|^2 $; \\
  } 
\end{algorithm}

At each step $\tau_{i}$, the denoising trajectory is updated to the previous time step $\tau_{i-1}$ by
\begin{equation}\label{DDIM trajectory}
\begin{split}
\mathbf{w}_{\tau_{i-1}}=\sqrt{\bar{\alpha}_{\tau_{i-1}}} \hat{\mathbf{w}}_0(\mathbf{w}_{\tau_i}, \tau_i, \mathbf{c}) + \sqrt{1-\bar{\alpha}_{\tau_{i-1}} } \hat{\boldsymbol{\epsilon}},
\end{split}
\end{equation}
where $\hat{\boldsymbol{\epsilon}}$ is computed based on Eq.~\eqref{noised precoding} using $\hat{\mathbf{w}}_0(\mathbf{w}_{\tau_i}, \tau_i, \mathbf{c})$, and is expressed as
\begin{equation}\label{pred noised}
\begin{split}
\hat{\boldsymbol{\epsilon}} = \frac{\mathbf{w}_{\tau_i}-\sqrt{\bar{\alpha}_{\tau_i}}  \hat{\mathbf{w}}_0(\mathbf{w}_{\tau_i}, \tau, \mathbf{c}) }{\sqrt{\left( 1-\bar{\alpha}_{\tau_i} \right)}}. 
\end{split}
\end{equation}

When the reverse denoising process of DDIM reaches the final time step, the output of the diffusion model is denoted by $\bar{\mathbf{w}}_0$, which will be unflattened into precoding matrices and serves as the action for $\mathcal{A}_{\mathcal{W}}$. $\mathcal{A}_{\mathcal{W}}$ together with the UAV selection action $\mathcal{A}_{\mathcal{U}}$ and the Top-K ratio action $\mathcal{A}_{\kappa}$ will jointly determine the reward in Eq.~\eqref{reward}. Specifically, $\mathcal{A}_{\mathcal{U}}$ and $\mathcal{A}_{\kappa}$ are determined by a $\mathcal{Q}$-network $\boldsymbol{\theta}_{\mathcal{Q}}$, which outputs the action-value $\mathcal{Q}(\cdot)$ for all candidate actions. The parameters of the network $\boldsymbol{\theta}_{\mathcal{Q}}$ are updated by minimizing the mean-squared error (MSE) loss, given by
\begin{equation}\label{loss_fucntion_DRL}
\begin{split}
\mathcal{L}(\boldsymbol{\theta}_{\mathcal{Q}})
= \frac{1}{2}  \left\| \mathcal{Q}(\mathcal{S}^t,\mathcal{A}^t_{\mathcal{U}}, \mathcal{A}_{\kappa}^t)- \mathcal{R} (\mathcal{S}_t, \mathcal{A}^t_{\mathcal{U}}, \mathcal{A}_{\kappa}^t, \mathcal{A}^t_{\mathcal{W}} )  \right\|^2. 
\end{split}
\end{equation}

The detailed procedures for the propsoed DDIM-based DRL algorithm are presented in \textbf{Algorithm}~\ref{algorithm1}. Compared with the conventional diffusion model, the computational complexity of DDIM is reduced from $\mathcal{O}(N_eT)$ to  $\mathcal{O}(N_e\Delta\tau)$, where $\Delta\tau =T/D$ denotes the sampling interval of the denoising process and $N_e$ is the number of DRL training epochs. The adoption of DDIM significantly improves computational efficiency while maintaining high-quality precoding performance.

\section{Simulation Results}

\begin{figure*}
\centering
\includegraphics[width=1.0\textwidth]{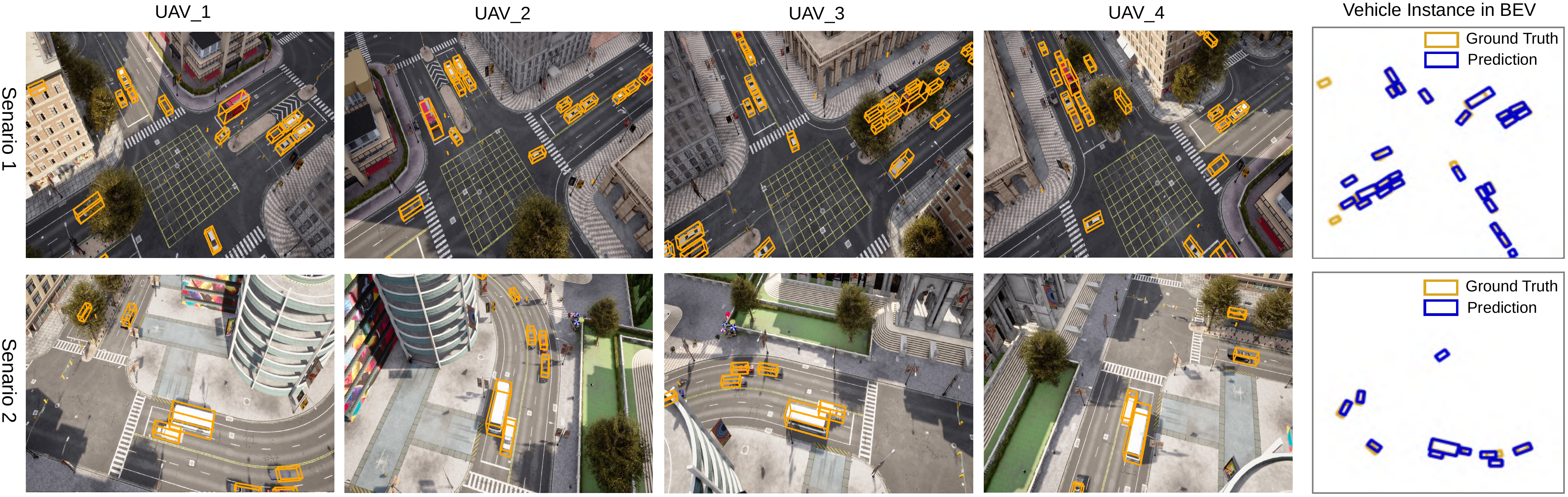} 
\centering
\caption{Visualization of 3D object detection results in the BEV representation.}
\label{figure_Visualization}
\end{figure*}

\begin{figure*}[htbp]
    \centering
    \begin{minipage}{0.325\textwidth}
        \centering
        \includegraphics[width=\linewidth]{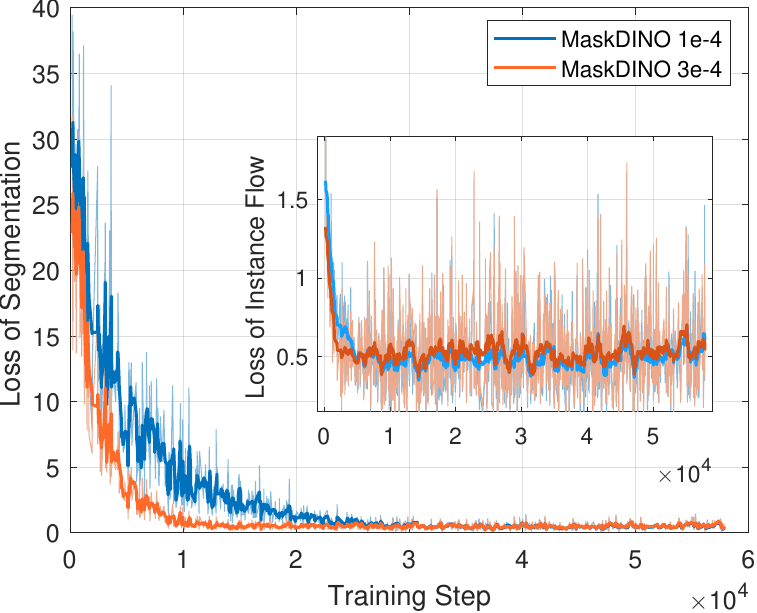}
        \captionof{figure}{ The training loss of segmentation and predicted instance flow.}
        \label{Simulation1}
    \end{minipage}
    \hfill
    \begin{minipage}{0.325\textwidth}
        \centering
        \includegraphics[width=\linewidth]{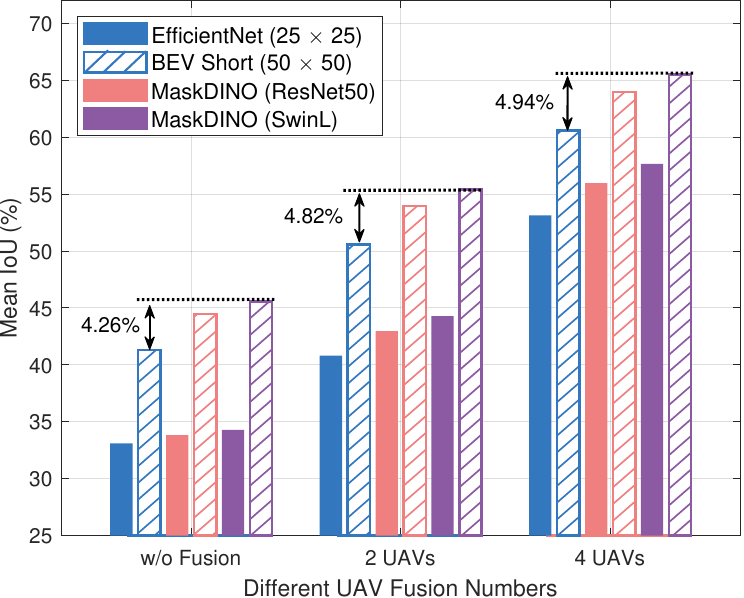} 
        \captionof{figure}{ Mean IoU of different perception backbones under varying UAV numbers.}
        \label{Fig_UAV_num}
    \end{minipage}
    \hfill
    \begin{minipage}{0.325\textwidth}
        \centering
        \includegraphics[width=\linewidth]{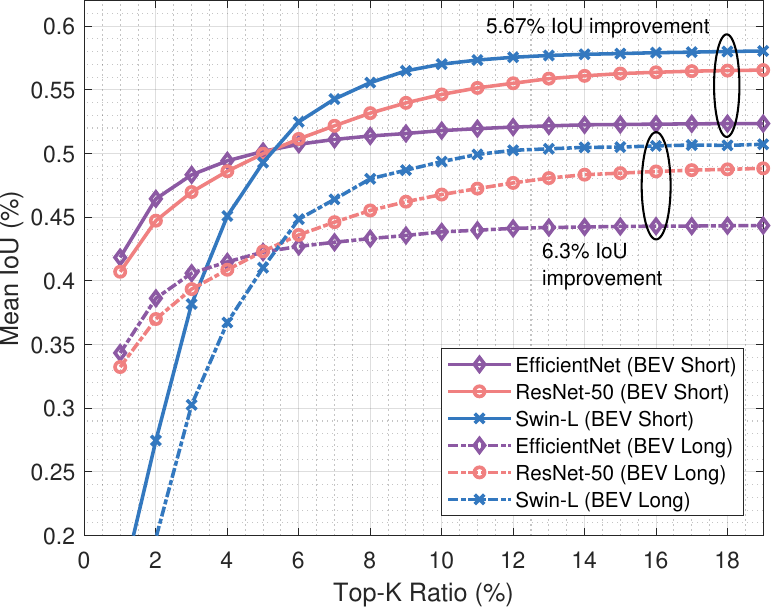}
        \captionof{figure}{Mean IoU versus Top-K ratios under BEV-Short and BEV-Long settings.}
        \label{Simulation3}
    \end{minipage}
\end{figure*}

This section evaluates the proposed BHU framework and DDIM-based DRL algorithm on the Air-Co-Pred \cite{wang2024drones} dataset. 
In Air-Co-Pred, $4$ UAVs hovering at an altitude of $50$ m monitor ground traffic flows from different viewpoints, covering an area of approximately $100$ m$\times$$100$ m. The dataset is generated using the CARLA simulator \cite{dosovitskiy2017carla}, providing $200$ realistic urban traffic scenes with a total of 32,000 synchronized images collected at a sampling rate of $2$ Hz. 
Each image has a resolution of $1600$ $\times$ $900$ pixels, and the dataset supports multi-frame temporal prediction tasks. Based on this dataset, we construct 5,780 training sequences and 1,020 testing sequences, where each sequence contains seven consecutive frames. Specifically, the first three frames are used as inputs, and the remaining four frames are used for prediction.

For the wireless communication links between the UAVs and the BS, the 3D channel model with a carrier frequency of 3.5 GHz is employed based on the 3GPP 38.901 \cite{3GPP38.901} standard. Each UAV is equipped with a $2$ $\times$ $1$ dual-polarized antenna array with $4$ antenna elements, while the BS employs a $2$ $\times$ $2$ dual-polarized antenna array with $8$ antenna elements. All selected UAVs share the same time–frequency resources and transmit data to the BS using OFDM symbols, which consists of $72$ subcarriers with a subcarrier spacing of $15$ kHz.

\subsection{Visualization and Performance of the BHU Scheme}
Fig.~\ref{figure_Visualization} presents two representative scenarios with 3D object detection visualizations, demonstrating vehicle instance annotations from different viewpoints of four UAVs. The rightmost column compares the predicted instances (shown in blue) with the ground-truth annotations (shown in gold) in the BEV plane. The substantial overlap between the predicted and ground-truth bounding boxes indicates the effectiveness and high accuracy of the proposed BHU method, even in challenging scenes with scattered or partially occluded vehicles.

Fig.~\ref{Simulation1} illustrates the training dynamics of MaskDINO under the learning rate settings of $1\times10^{-4}$ and $3 \times 10^{-4}$ for both the segmentation loss and the instance flow loss. The configuration with a learning rate of $3 \times 10^{-4}$ exhibits significantly faster convergence in the early training stage, whereas the learning rate of $1 \times 10^{-4}$ yields smoother training trajectories while achieving comparable final convergence performance. These results indicate that MaskDINO can be trained reliably to convergence in different learning rate settings, and that selecting a learning rate of $3 \times 10^{-4}$ provides a reasonable trade-off between convergence speed and training stability.

\begin{table*}[t]
\centering
\vspace{-4mm}
\caption{ Performance comparison of EfficientNet- and MaskDINO-based networks with/without BEV fusion.}
\resizebox{0.95\textwidth}{!}{
\begin{tabular}{lccccccc}
\toprule
\rowcolor{gray!08}
{\textbf{Method}} & \multicolumn{3}{c}{\textbf{IoU (\%) ↑}} & \multicolumn{3}{c}{\textbf{PQ (\%) ↑}} & \textbf{Comm. Cost (Bps) ↓} \\
\cmidrule(lr){2-4} \cmidrule(lr){5-7}
DHD ~\cite{wang2024drones} & Short & Long & Avg. & Short & Long & Avg.  & Top-k (25\%) \\
\midrule
EfficientNet (4 UAVs) & 60.58 & 53.14 & 56.86 & 51.80  & 45.89 & 48.85  & $ 6.14 \times 10^7 $ \\
EfficientNet (2 UAVs) & 50.59  & 40.82 & 45.71 & 43.43 & 35.37 & 39.40 &  $  2.05 \times 10^7 $  \\
EfficientNet w/o Fusion & 41.31 & 33.11 & 37.21 & 35.07 & 28.29 & 31.68 & $ 0 $ \\
\midrule
\textbf{MaskDINO} (ResNet50) & 63.99 & 55.97 & 59.98 & \textbf{54.24} & 48.23 & 51.24 & $ 6.14 \times 10^7 $ \\
\textbf{MaskDINO} (SwinL) & \textbf{65.52} & \textbf{57.66} & \textbf{61.59}  & 54.10 & \textbf{50.63} & \textbf{52.37}& $ 6.14 \times 10^7 $ \\
\textbf{MaskDINO} (SwinL) w/o Fusion & 45.57 & 34.30 & 39.94  & 39.50 & 30.26 & 34.88 & $ 0$ \\
\bottomrule
\end{tabular}
}
\label{TableI}
\end{table*}

\begin{table}[t]
\centering
\caption{ Results of backbone networks and LoRA fine-tuning.}
\resizebox{0.5\textwidth}{!}{
\begin{tabular}{lcccc}
\toprule
\rowcolor{gray!08}
{\textbf{Method}} & \multicolumn{2}{c}{\textbf{Metrics (\%) ↑}} &  \multicolumn{2}{c}{\textbf{Parameters ↓}} \\
 \cmidrule(lr){2-3} \cmidrule(lr){4-5}
SwinL (Backbone)  & IoU & Prec.  & Total & Training\\
\midrule
EfficientNet \textbf{Baseline} & 52.89 & 70.15  & \textbf{39.5} M  & 39.5 M  \\
BHU w pretrained & \textbf{57.81}  & \textbf{77.57}  & 0.3 B  & 0.3 B\\
BHU w/o pretrained  & 53.27 & 72.22  & 0.3 B & 0.3 B\\
\midrule
\textbf{DHD} + \textbf{LoRA} & 55.24  &  73.40  & $\sim$0.3 B & \textbf{37.4} M \\
\textbf{BHU} + \textbf{LoRA} & 55.86 & 74.38  & $\sim$0.3 B & \textbf{37.4} M \\
\bottomrule
\end{tabular}
}
\label{TableII}
\end{table}

\begin{figure}
\centering
\includegraphics [width=0.46\textwidth]{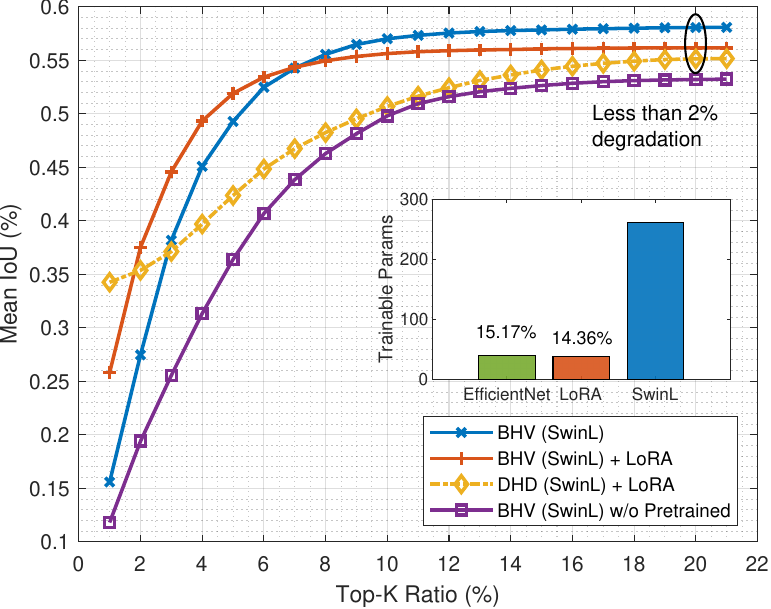} 
\captionsetup{justification=justified,format=plain}
\caption{ IoU versus Top-K ratios under SwinL and LoRA settings.
}
\label{Simulation4}
\end{figure}

To evaluate the performance of the proposed BHU scheme under different network backbones and varying numbers of UAVs, Fig.~\ref{Fig_UAV_num} compares the mean IoU in a long BEV setting of $50 \times 50$ and a short setting of $25 \times 25$. 
The evaluated backbones include a CNN-based EfficientNet, a ResNet-50-based lightweight MaskDINO, and a Swin-large-based MaskDINO. The simulation results show that increasing the number of cooperative UAVs consistently improves the mean IoU across all considered backbones and BEV configurations, indicating that multi-view observations provide complementary visual information that effectively enhances perception accuracy. In particular, the LVM-based backbone yields significant performance gains for all 1, 2, and 4 UAV configurations, achieving relative improvements ranging from 4.26\% to 4.94\% compared to the EfficientNet-based baseline.

Fig.~\ref{Simulation3} depicts the mean IoU performance versus different Top-K transmission ratio under both the long BEV and short BEV settings. For all evaluated backbones, increasing the Top-K ratio leads to improved perception accuracy, indicating that re3etaining a larger portion of images helps preserve geometric feature information.
However, the IoU improvement gradually saturates when the sparsification ratio exceeds approximately 15\%, which is attributed to the fact that the perception task focuses on vehicle detection and a substantial portion of background information can be discarded.
Moreover, as highlighted in Fig.~\ref{Simulation3}, the Swin-large achieves 5.67\% and 6.3\% larger relative IoU gains compared to the EfficientNet under the BEV-short and BEV-long settings, respectively. The simulation results demonstrate the enhanced model generalization and representation capacity of LVMs due to their increased network depth advanced architectural designs.

\subsection{Comparison with the DHD Baseline}
We further conduct benchmark comparisons in terms of collaborative perception utility and communication overhead with a representative Drone-Help-Drone (DHD) scheme~\cite{wang2024drones}. In DHD, a lightweight EfficientNet is deployed onboard to extract BEV features, and sparse BEV features from neighboring UAVs will be collected and fused for cooperative perception.

\begin{table*}[t]
\centering
\vspace{-4mm}
\caption{ Comparison of the proposed BHU scheme and baselines in terms of utility and communication cost.}
\resizebox{0.90\textwidth}{!}{
\begin{tabular}{lccccccc}
\toprule
\rowcolor{gray!08}
{\textbf{Method}} & \multicolumn{3}{c}{\textbf{IoU (\%) ↑}} & \multicolumn{3}{c}{\textbf{PQ (\%) ↑}} & \textbf{Comm. Cost (Bps) ↓} \\
\cmidrule(lr){2-4} \cmidrule(lr){5-7}
BHU (Proposed) & Short & Long & Avg. & Short & Long & Avg.  & Top-k (25\%) \\
\midrule
DHD (EfficientNet) \textbf{Baseline} & 60.58 & 53.14 & 56.86 & 51.80 & 45.89 & 48.85  &    6.14 $\times$  $10^7 $   \\
BHU (EfficientNet) & 57.39 & 52.89 & 55.14 & 47.23 & 45.68 & 46.46 &  $ \textbf{9.22} \times 10^6 $  \\
\midrule
\textbf{MaskDINO} (ResNet50) & 64.00 & 56.32 & 60.16 & \textbf{56.05} & 49.40 & 52.73 & $ \textbf{9.22} \times 10^6 $ \\
\textbf{MaskDINO} (SwinL) & \textbf{65.71} & \textbf{57.81} & \textbf{61.76}  & 54.64 & \textbf{51.26} & \textbf{52.95}& $ \textbf{9.22} \times 10^6 $ \\
\textbf{MaskDINO} (SwinL) + \textbf{LoRA}  & 63.51 & 55.86 & 59.69  & 53.74 & 48.58 & 51.16 & $\textbf{9.22} \times 10^6$  \\
\bottomrule
\end{tabular}
}
\label{TableIII}
\end{table*}

\begin{figure*}[h]
    \centering
    \begin{minipage}{0.42\textwidth}
        \centering
        \includegraphics[width=\linewidth]{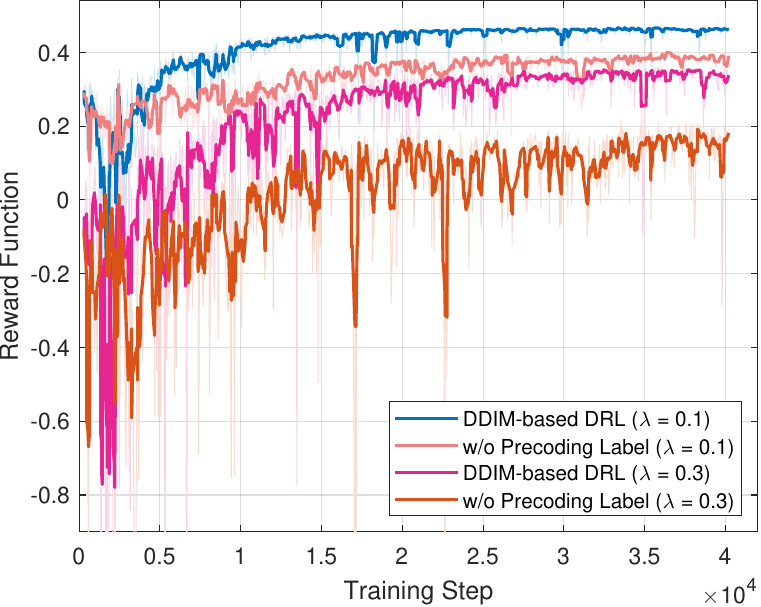}
        \captionof{figure}{Training reward comparison with and without search-based precoding labels.}
        \label{Simulation5}
    \end{minipage}
    \hspace{0.05\textwidth} 
    \begin{minipage}{0.42\textwidth}
        \centering
        \includegraphics[width=\linewidth]{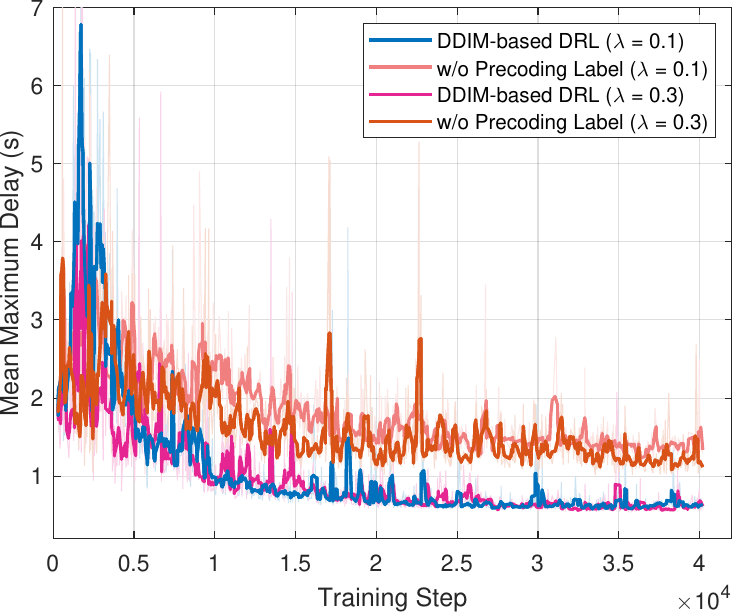} 
        \captionof{figure}{ Training latency comparison with and without search-based precoding labels.}
        \label{Simulation6}
    \end{minipage}

    \begin{minipage}{0.43\textwidth}
        \centering
        \includegraphics[width=\linewidth]{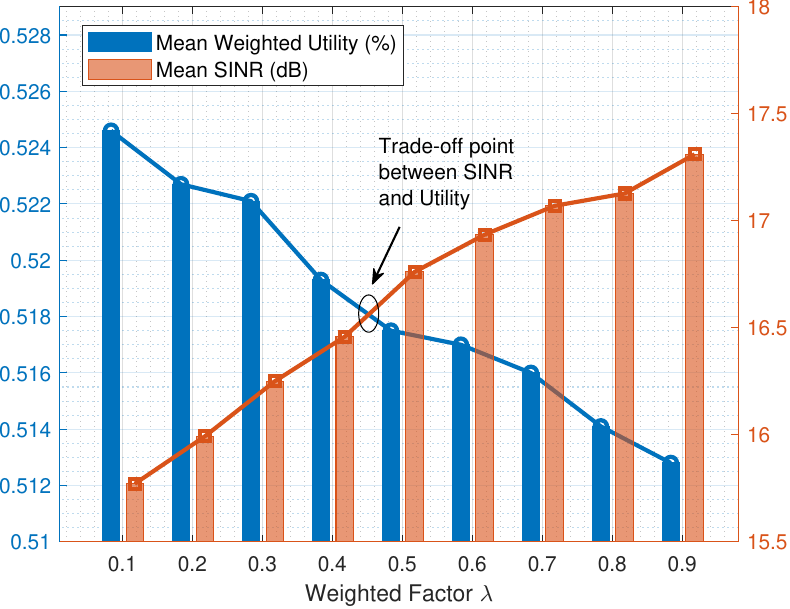}
        \captionof{figure}{ Trade-off between utility and SINR under different weighted factors $\lambda$.}
        \label{Simulation7}
    \end{minipage}
    \hspace{0.025\textwidth} 
    \begin{minipage}{0.42\textwidth}
        \centering
        \includegraphics[width=\linewidth]{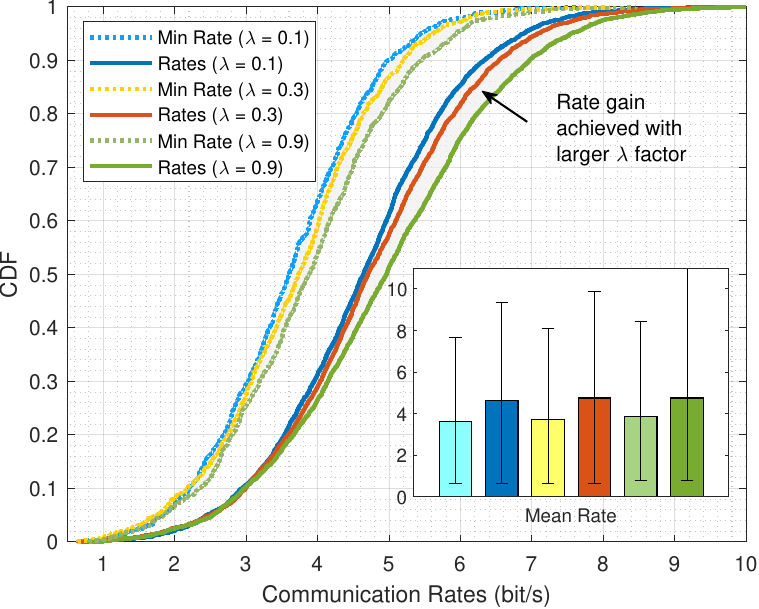} 
        \captionof{figure}{ Communication rates of UAVs under different weighted $\lambda$ factors.}
        \label{Simulation9}
    \end{minipage}

\end{figure*}

As reported in Table~\ref{TableI}, similar to the BHU scheme, increasing the number of cooperative UAVs consistently improves both IoU and PQ performance under all BEV configurations. On the other hand, involving more UAVs in cooperation inevitably incurs additional communication overhead. Specifically, for the DHD scheme without feature fusion, no visual data are exchanged and hence the communication overhead is zero. When $2$ or $4$ UAVs participate in cooperative perception, assuming a Top-k ratio of 25\% and a payload of 2 bytes per pixel, the communication overhead of transmitted BEV features (64$\times$200$\times$200) increases to $2.05 \times 10^7 $ and $2.05 \times 10^7 $ Bps, respectively.
Additionally, when the backbone of EfficientNet is replaced with a Swin-large-based MaskDINO, the average IoU and PQ are improved from 56.86\% and 48.85\% to 61.59\% and 52.37\%, which exhibits the enhanced capability of LVMs under the same communication overhead.

Table~\ref{TableII} further summarizes the scale impact of the network backbone on perception performance. Compared to EfficientNet, the Swin-large model, which contains approximately 0.3 B parameters (about 7.6$\times$ more than EfficientNet), achieves improvements of 4.92\% in mean IoU and 7.42\% in prediction accuracy. However, retraining Swin-large without pretrained weights yields only marginal performance gains and incurs significant training overhead. Additionally, we investigate the application of low-rank adaptation (LoRA) fine-tuning, which introduces only 37.4 M (14.36\%) trainable parameters to the Swin-large backbone. As illustrated in Fig.~\ref{Simulation4}, the Swin-large with LoRA setting exhibits less than a 2\% degradation in IoU compared to the full training with pretrained weights, requiring even fewer trainable parameters than EfficientNet (15.17\%).

Unlike the DHD scheme, the proposed BHU framework transmits Top-K RGB images and performs BEV fusion using the Swin-large backbone at a ground server. As presented in Table~\ref{TableIII}, the communication overhead of BHU is measured at $9.22 \times 10^6 $ bps when the input image resolution is resized to $224 \times 480$ and the Top-K ratio is set to 25\%, which accounts for only 15\% of the overhead required by the DHD benchmark. Moreover, by leveraging the representation capability of LVMs, BHU achieves average IoU and PQ gains of 4.90\% and 4.10\% over DHD. Under the LoRA fine-tuning setting, the average IoU and PQ are still improved by 2.83\% and 2.31\%, respectively. These results demonstrate that BHU provides an effective solution to reduce communication overhead while enhancing cooperative perception performance among UAVs.

\subsection{Performance of DDIM-based DRL Algorithm }

Based on the BHU performance under different numbers of cooperative UAVs and Top-K ratios, we evaluate the effectiveness of the proposed DDIM-based DRL algorithm in jointly optimizing communication rates and perception utility. 


Fig.~\ref{Simulation5} illustrates the training reward of the proposed DRL approach under latency weight factors $\lambda = 0.1$ and $\lambda = 0.3$. A higher reward indicates that the DRL policy achieves better perception performance in terms of IoU and PQ, as well as higher wireless transmission efficiency. It can be observed that, for both $\lambda$ settings, the incorporation of DDIM module for modeling the distribution of optimal precoding labels (obtained through exhaustive codebook search) leads to faster and more stable convergence. In contrast, the scheme without optimal precoding labels exhibits larger fluctuations in the training reward and converges to a relatively lower reward value, demonstrating the effectiveness of DDIM in generating precoding actions for the DRL training process.


In addition, since the reward function is formulated as the difference between the perception utility and the communication latency penalty, a larger latency weight ($\lambda$ = 0.3) places greater emphasis on the communication penalty. As a result, the overall reward value is lower than that obtained under the setting of $\lambda$ = 0.1. 
Fig.~\ref{Simulation6} illustrates the reduction and convergence behavior of UAV communication latency during the DRL training process, validating the efficacy of the designed reward function in improving transmission efficiency. Specifically, for $\lambda$ = 0.1, the DDIM-based approach achieves a lower latency than the scheme without optimal precoding labels, even when compared with the latter under a higher latency weight of $\lambda$ = 0.3. These simulation results further verify the advantage of DDIM-generated precoding in enhancing communication performance for multi-UAV scenarios.


Subsequently, Fig.~\ref{Simulation7} illustrates the variation curves of the weighted perception utility and the average signal-to-noise ratio (SNR) as the latency weight factor $\lambda$ increases from 0.1 to 0.9.
It can be observed that the average SNR of cooperative UAVs increases as the latency weight factor $\lambda$ increases, whereas the average perception utility gradually decreases. This behavior is attributed to the role of $\lambda$ in regulating the trade-off between perception utility and communication latency, which can adjust the penalty imposed on latency during the DRL training process. In addition, Fig.~\ref{Simulation7} reveals a turning point characterized by the intersection of the perception utility and SNR curves, which provides practical guidance for selecting an appropriate weighting factor to balance the perception and communication performance.


Fig.~\ref{Simulation9} presents the cumulative distribution function (CDF) of the minimum and total communication rates of the cooperative UAVs under different $\lambda$ settings. The inset illustrates the average communication rate along with the corresponding standard deviation. Similar to the SNR performance in Fig.~\ref{Simulation7}, both the minimum and total communication rates increase with the value of $\lambda$.
Consequently, under a higher $\lambda$ setting, the DDIM-based DRL algorithm tends to select UAVs with more favorable channel conditions, along with more optimized Top-K transmission and precoding strategies, thereby providing a communication-efficient strategy under resource-constrained wireless environments.







\section{Conclusion}

In this paper, we have investigated a communication-aware cooperative perception framework for multi-UAV systems in low-altitude economy scenarios, aiming to address the critical challenges of excessive visual data transmission and stringent latency constraints. Specifically,
we have proposed a Top-K sparsification mechanism to enable efficient visual transmission by selecting the most informative pixels from UAV-captured images. An LVM–based perception pipeline deployed at the ground server has been established to perform BEV feature extraction and feature fusion based on the sparsified visual data. Furthermore, a DDIM-based DRL algorithm was developed to jointly optimize cooperative UAV selection, sparsification ratios, and precoding matrices, achieving an effective trade-off between perception utility and communication efficiency. In future work, we will extend the proposed framework to more dynamic UAV mobility scenarios and integrate the advanced reasoning capabilities of large foundation models into intelligent applications in low-altitude economy networks.

\bibliographystyle{IEEEtran}
\bibliography{Ref}


\end{document}